\documentclass[10pt,twocolumn,letterpaper]{article}

\usepackage{cvpr}              
\definecolor{cvprblue}{rgb}{0.21,0.49,0.74}
\usepackage[pagebackref,breaklinks,colorlinks,allcolors=cvprblue]{hyperref}

\usepackage{graphicx}
\usepackage{booktabs}
\usepackage{amsfonts}
\usepackage{amsmath}
\usepackage{pifont}
\usepackage{multirow}
\usepackage{subcaption}
\usepackage{enumitem}
\usepackage{tabularx}
\usepackage{makecell}
\newcolumntype{Y}{>{\centering\arraybackslash}X}

\newcommand*{\netname}{SFSNet}
\newcommand*{\modname}{SBF}
\newcommand*{\schname}{SBFConv3D}

\def\paperID{38} 
\def\confName{ABAW}
\def\confYear{2026}

\title{SBF: An Effective Representation to Augment Skeleton for \\
Video-based Human Action Recognition}

\author{Zhuoxuan~Peng$^{1}$ 
\hspace{1em}
Yiyi~Ding$^{2}$ 
\hspace{1em}
Yang~Lin$^{1}$ 
\hspace{1em}
S.-H.~Gary~Chan$^{1}$\\
$^{1}$ The Hong Kong University of Science and Technology\\
$^{2}$ The Hong Kong University of Science and Technology (Guangzhou)
\\
{\tt\small zpengac@cse.ust.hk, ydingaz@connect.hkust-gz.edu.cn, \{lyangbe, gchan\}@cse.ust.hk}
}

\begin{document}
\maketitle
\begin{abstract}
Many modern video-based human action recognition (HAR) approaches use 2D skeleton as the intermediate representation in their prediction pipelines.
Despite overall encouraging results, these approaches still struggle in many common scenes, mainly because the skeleton  
misses
critical action-related information pertaining to the depth of the joints, contour of the human body, and interaction between the human and objects.  
To address this, we augment skeleton with a  novel and effective representation that captures action-related information 
in the pipeline of HAR without any extra annotation overhead beyond the existing skeleton extraction.
The representation, termed Scale-Body-Flow (\modname{}), consists of three distinct components, namely a map volume given by the scale (and hence depth information) of each joint, a body map outlining the human subject, and a flow map given by pixel-wise optical flow values due to human-object interaction.
To predict \modname{}, we further present \netname{}, a novel \textbf{s}egmentation \textbf{net}work supervised by the optical \textbf{f}low 
and \textbf{s}keleton. 
Extensive experiments across different datasets demonstrate that our pipeline based on \modname{} and \netname{} achieves significantly higher HAR accuracy with similar compactness and efficiency as compared with the state-of-the-art skeleton-only approaches.
\end{abstract}    
\begin{figure}[t]
    \centering
    \includegraphics[width=0.95\linewidth]{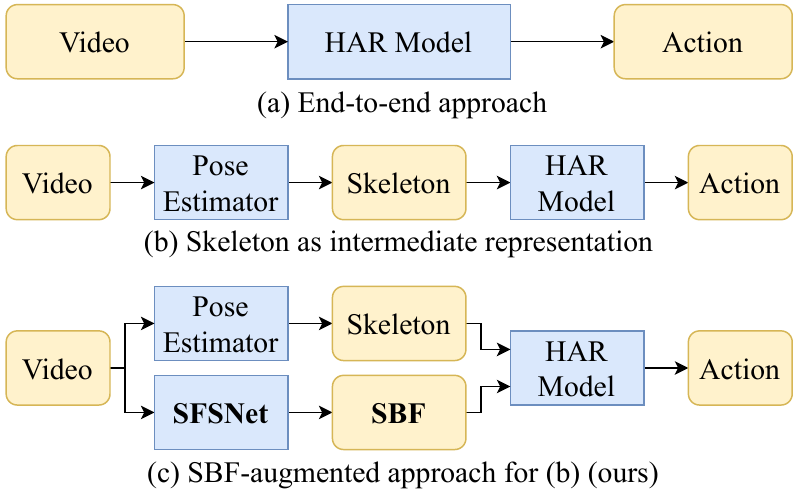}
    \caption{Comparison of video-based HAR pipelines. Our proposed pipeline (c) employs \modname{} predicted by \netname{} to augment skeleton for effective HAR, addressing the limitations of the skeleton-only approach (b).}
    \label{fig:teaser}
\end{figure}

\begin{figure}[t]
    \centering
    \begin{subfigure}{.27\linewidth}
        \centering
        \includegraphics[width=\linewidth]{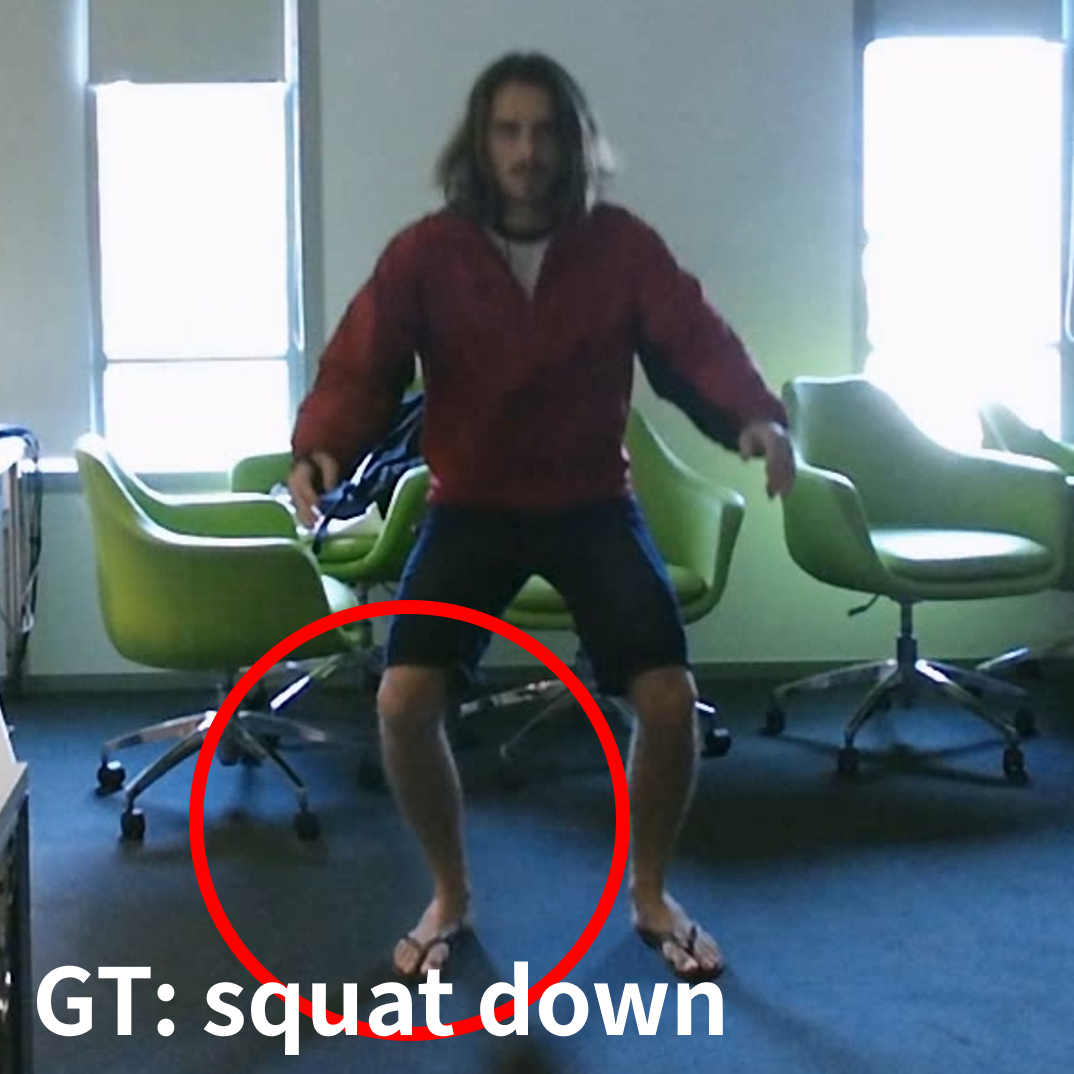}
    \end{subfigure}
    \begin{subfigure}{.27\linewidth}
        \centering
        \includegraphics[width=\linewidth]{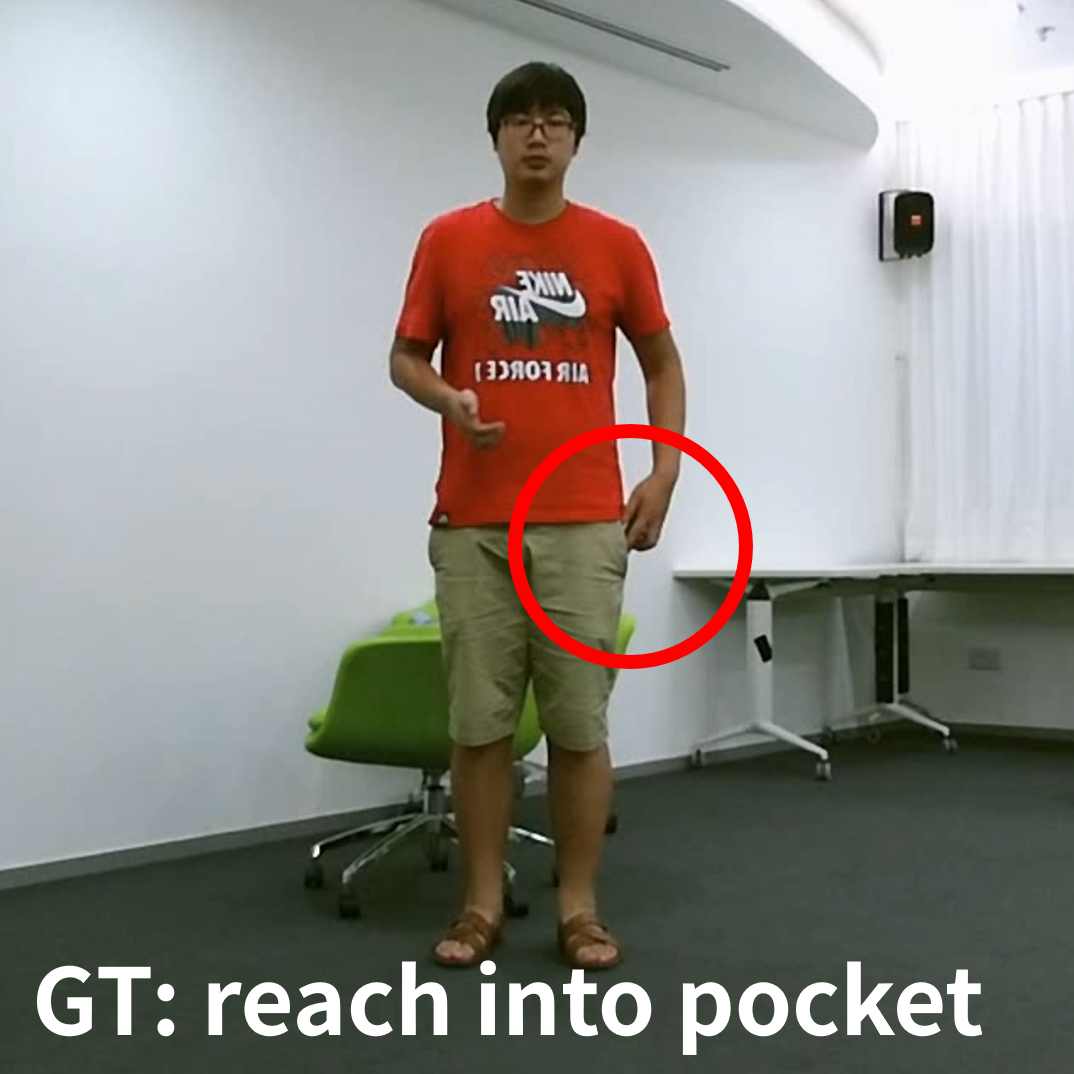}
    \end{subfigure}
    \begin{subfigure}{.27\linewidth}
        \centering
        \includegraphics[width=\linewidth]{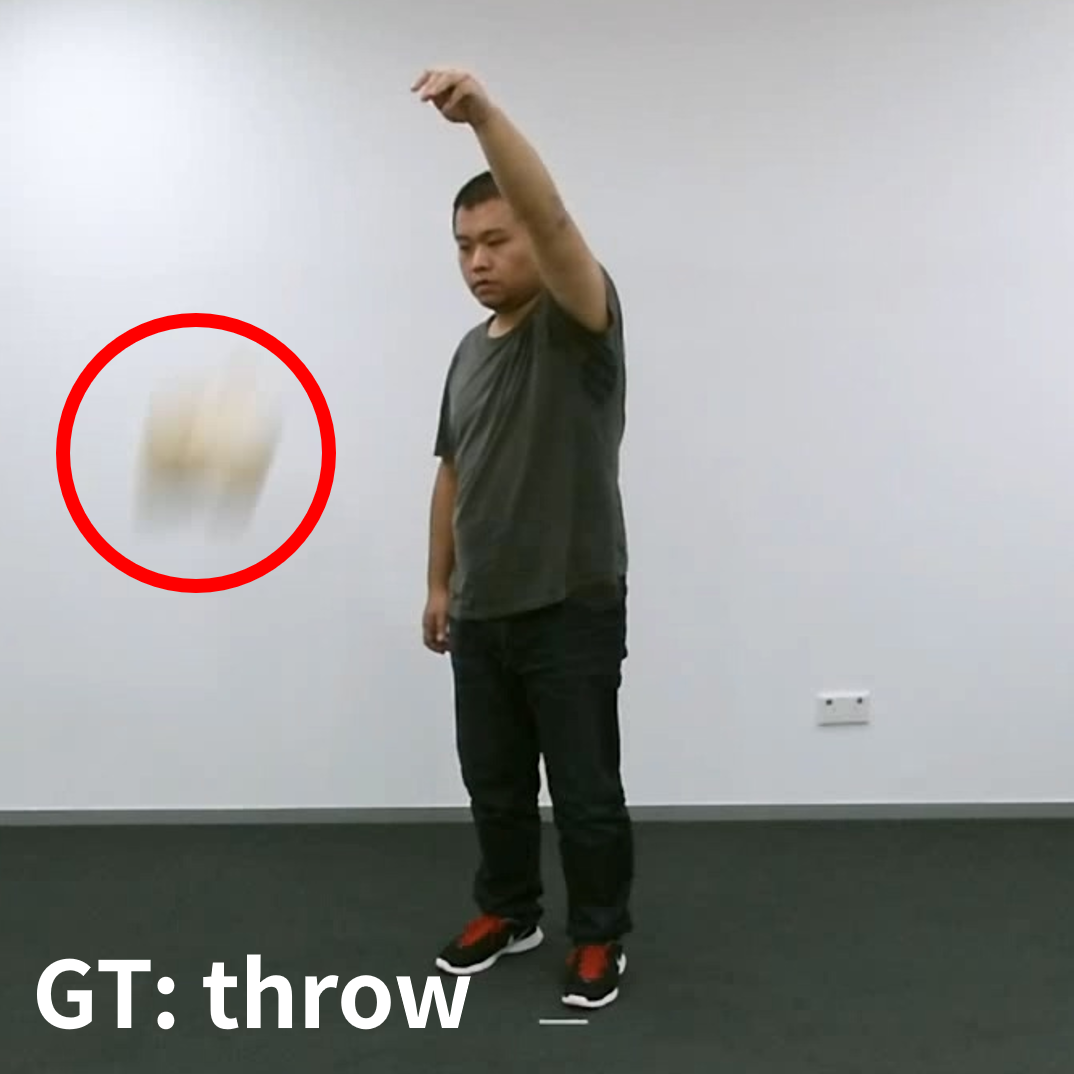}
    \end{subfigure}
    \begin{subfigure}{.27\linewidth}
        \centering
        \includegraphics[width=\linewidth]{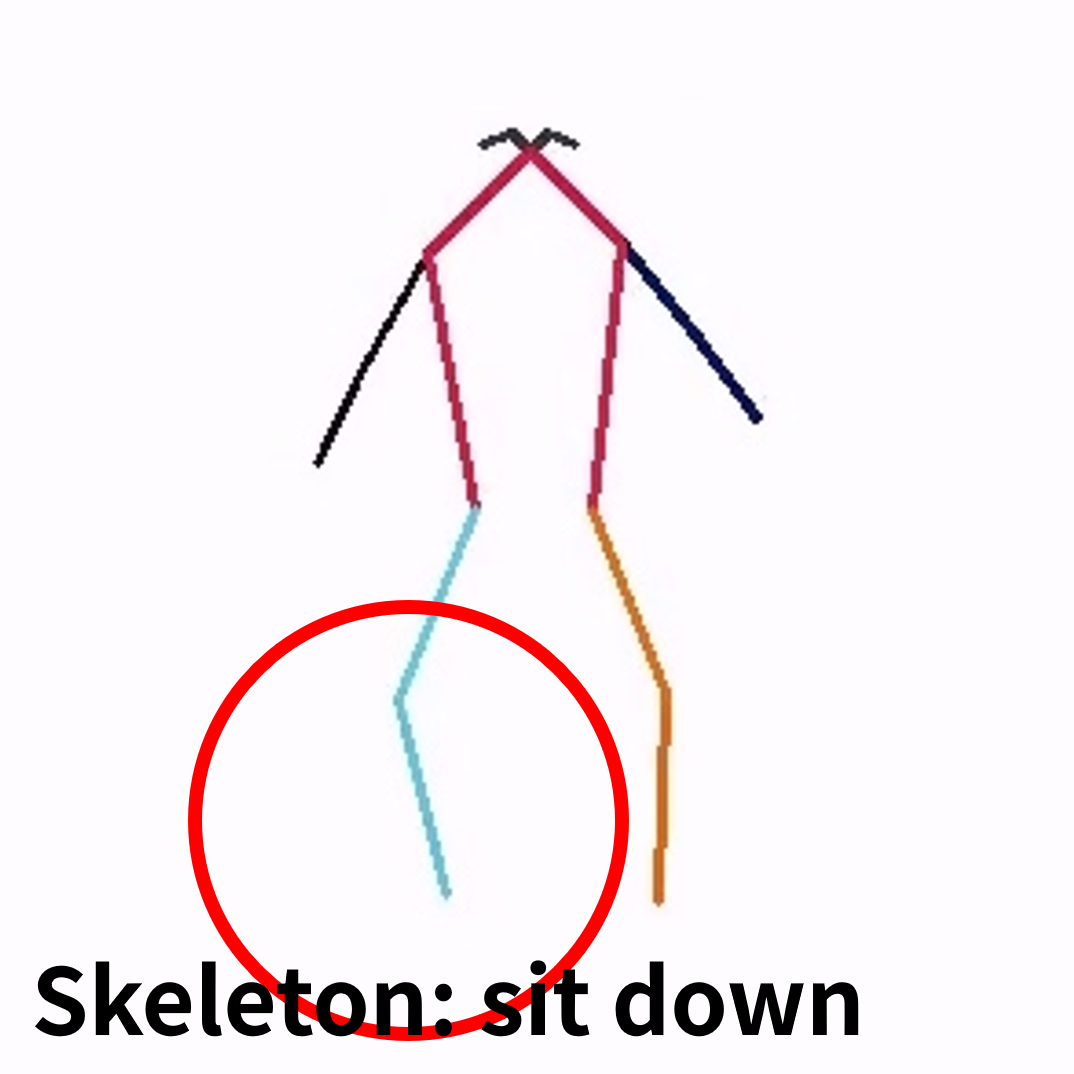}
    \end{subfigure}
    \begin{subfigure}{.27\linewidth}
        \centering
        \includegraphics[width=\linewidth]{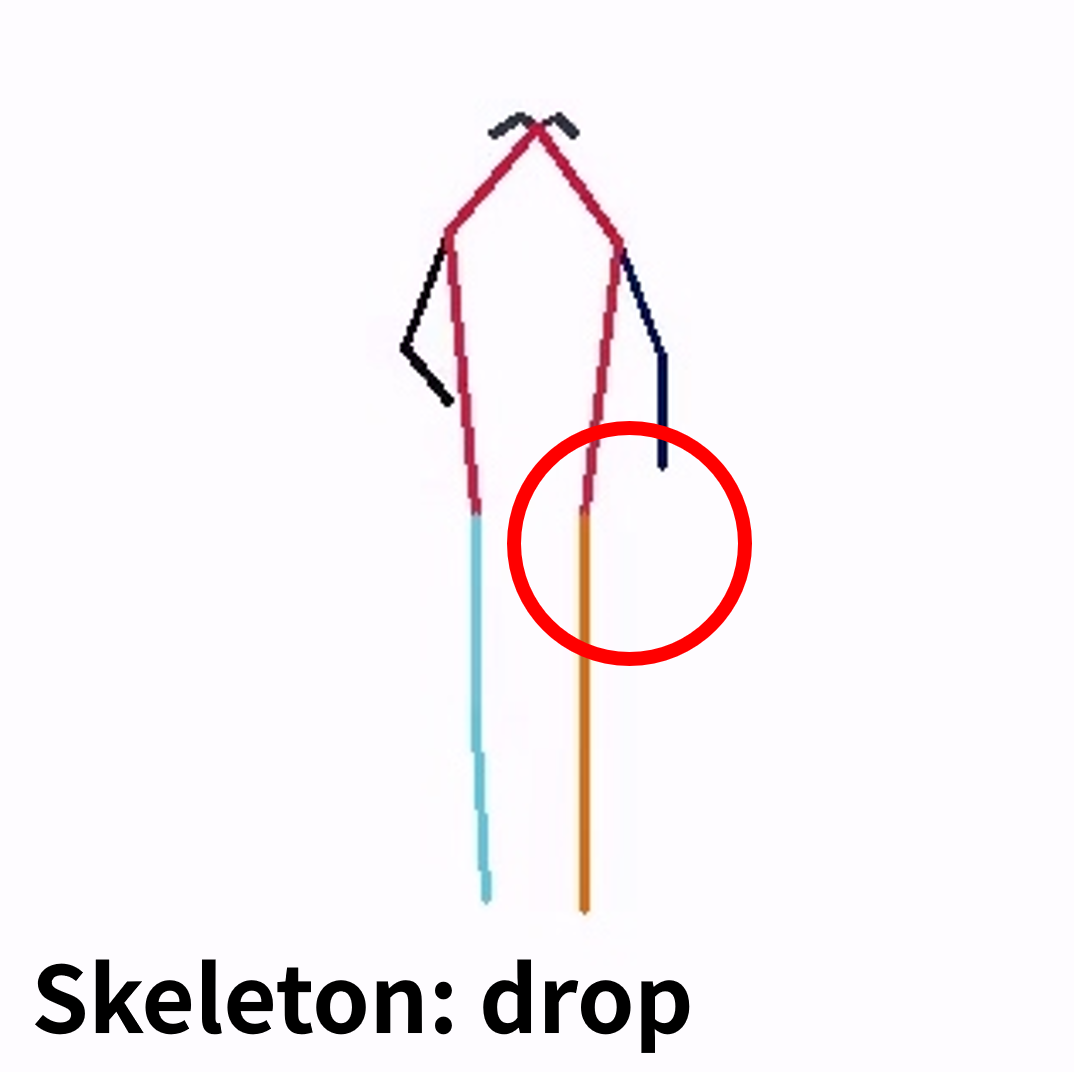}
    \end{subfigure}
    \begin{subfigure}{.27\linewidth}
        \centering
        \includegraphics[width=\linewidth]{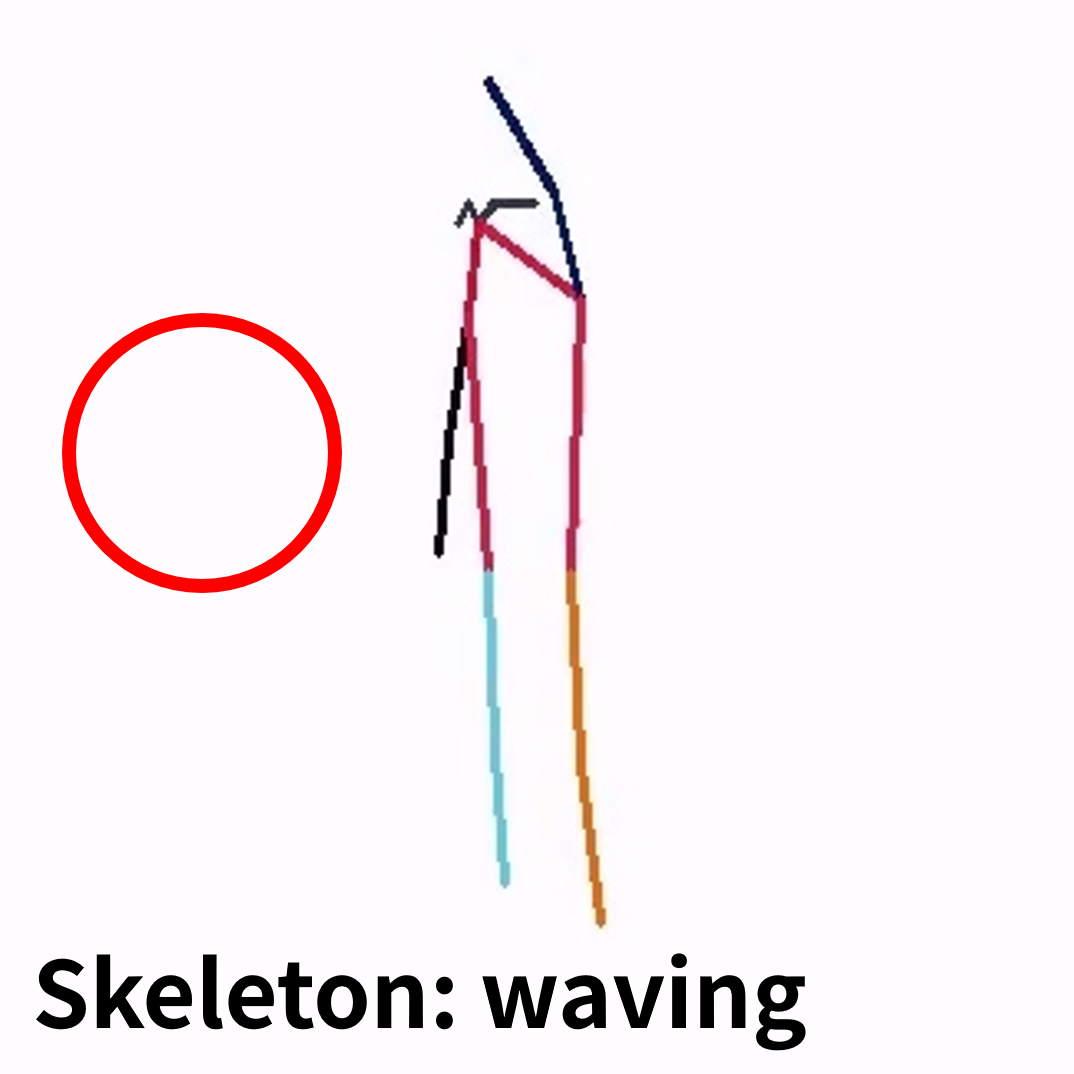}
    \end{subfigure}
    \begin{subfigure}[t]{.27\linewidth}
        \centering
        \includegraphics[width=\linewidth]{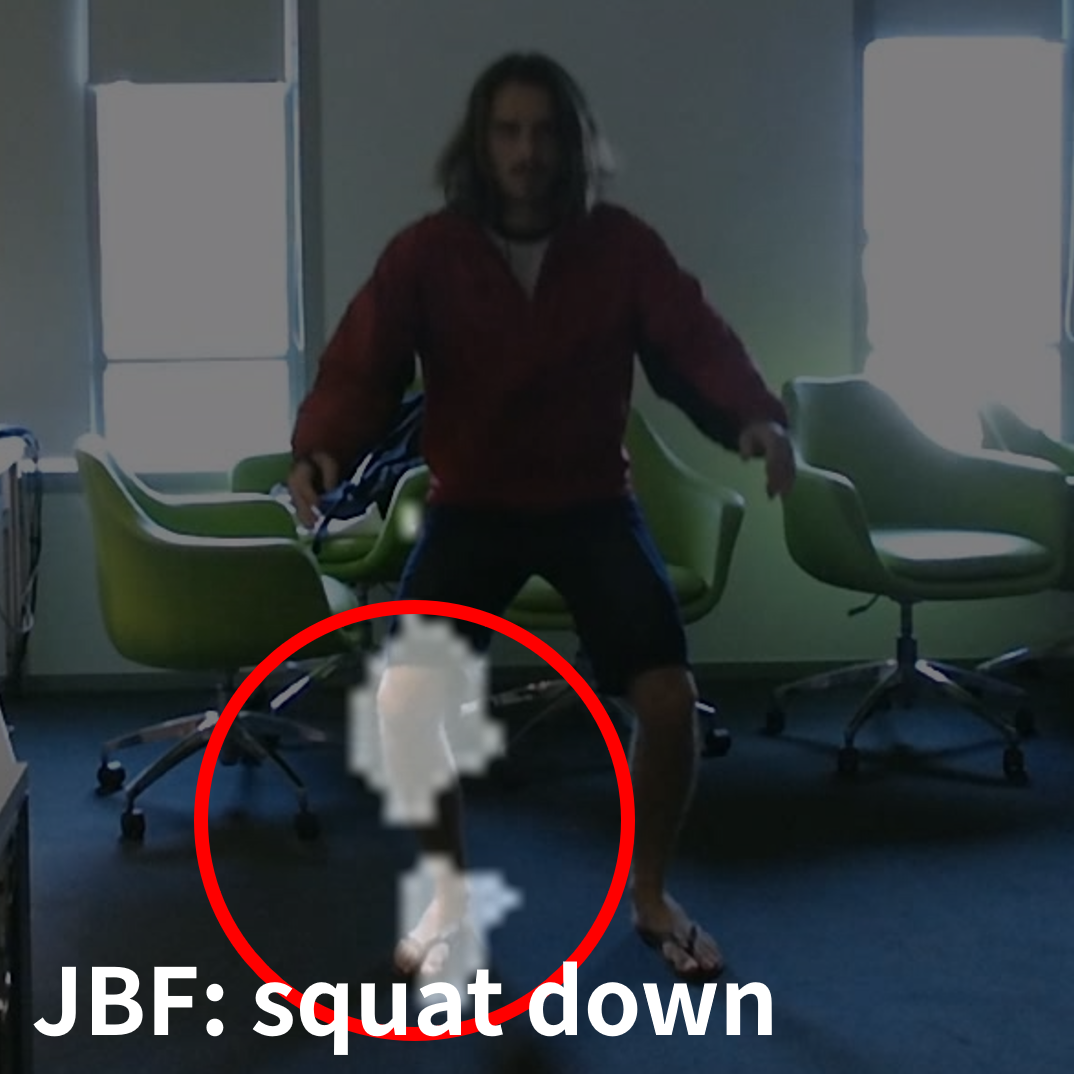}
        \caption{Joint depth}
    \end{subfigure}
    \begin{subfigure}[t]{.27\linewidth}
        \centering
        \includegraphics[width=\linewidth]{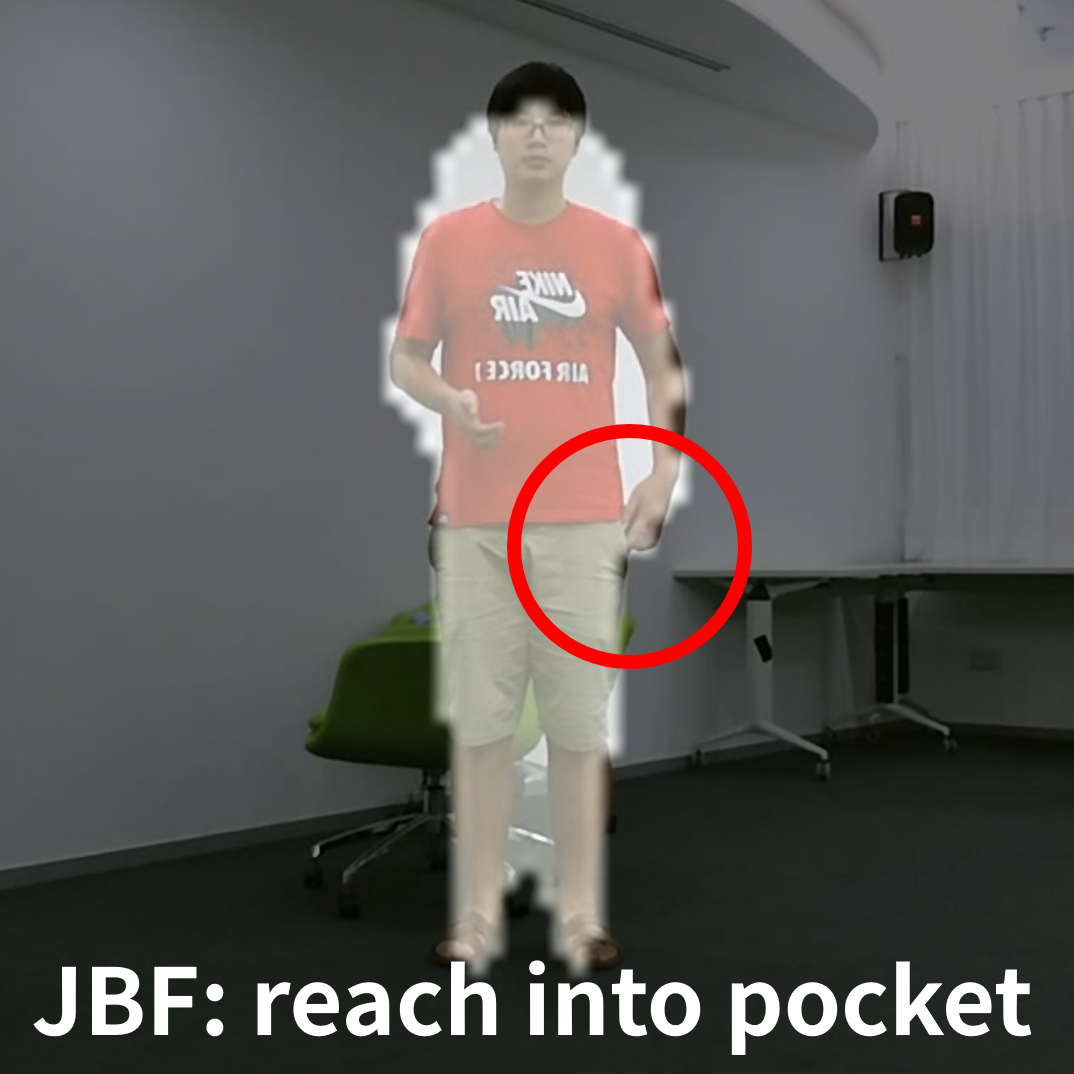}
        \caption{Body contour}
    \end{subfigure}
    \begin{subfigure}[t]{.27\linewidth}
        \centering
        \includegraphics[width=\linewidth]{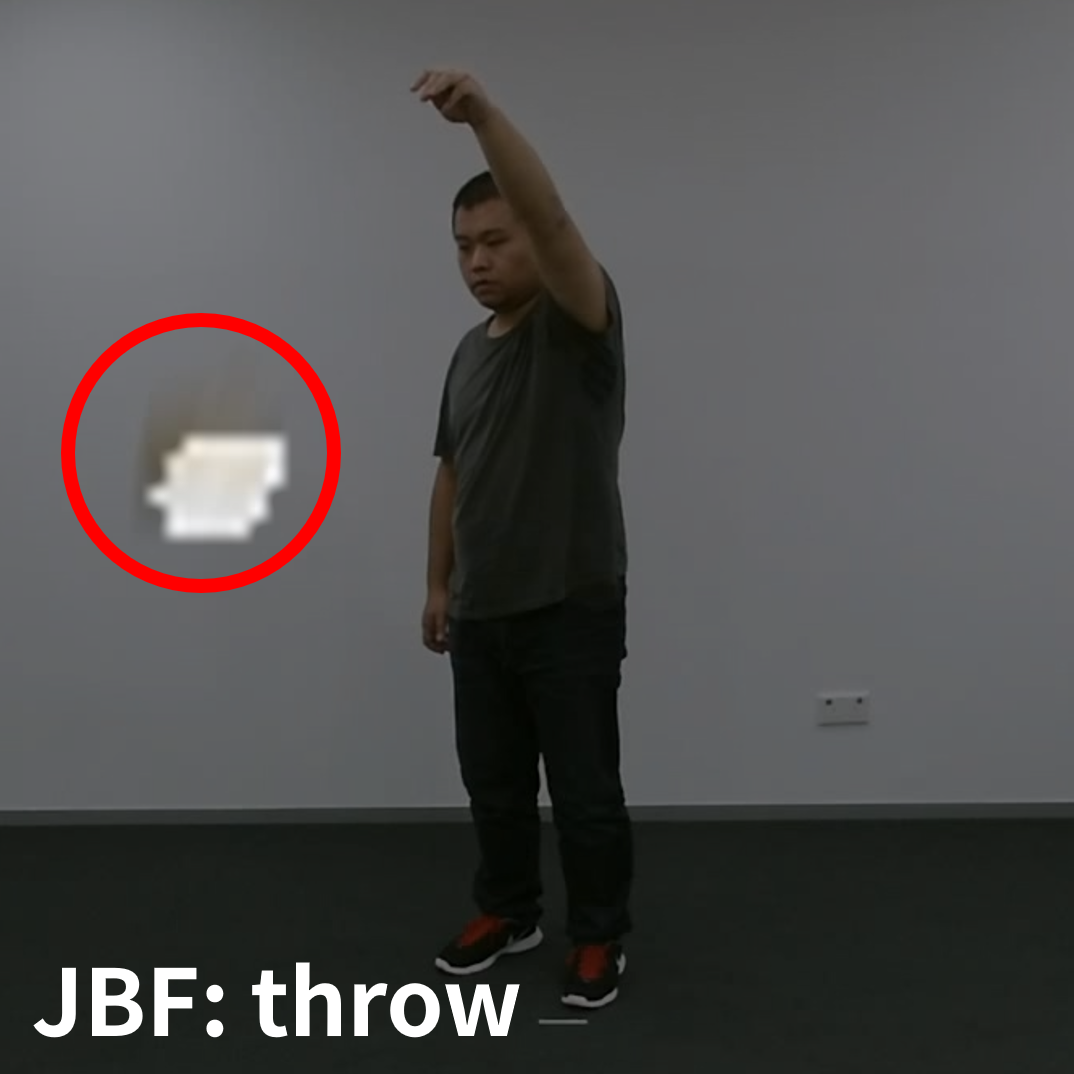}
        \caption{HOI}
    \end{subfigure}

    \caption{Failure cases of HAR based on extracting 2D skeletons from videos, addressed with our \modname{} representation predicted via \netname{}. Top row: original video frame; middle row: skeleton with prediction error; bottom row: correct prediction using skeleton+\modname{}, owing to the capturing of the action-related information. Red circles highlight key regions possibly leading to the error. 
    (a) \textbf{Joint depth:} The 2D skeleton's flat nature causes the ambiguity of the depth of each joint, e.g., confusing ``squatting down'' and ``sitting down'' from a front view. 
    (b) \textbf{Body contour:} The contour of human body, an encircling of all body parts, provides richer features than skeleton. E.g., ``reaching into a pocket'' is mis-predicted as ``drop'' because the subject's left hand is not shown overlapping with his body in the skeleton. 
    (c) \textbf{Human-object interaction (HOI):} The skeleton misses interactions between the subject and objects, e.g., ``throwing'' misinterpreted as ``waving'' due to the ignored motion between the subject and the thrown object.}
    \label{fig:skl_fail}
\end{figure}

\section{Introduction}
\label{sec:intro}

Video-based human action recognition (HAR) is to classify human actions from a video sequence. It has wide applications in robotics, healthcare, surveillance, and human-computer interaction. 
The most intuitive approach for HAR is direct end-to-end classification from input frames. Despite its impressive performance, it is too computationally intensive (due to the large model structure and volume of video pixels involved) to be practically deployed.

To reduce computational overhead, HAR based on intermediate representation has been adopted, where an action representation is first extracted from the video frames and then processed by a downstream network. Among various representations studied, the 2D human skeleton has emerged as the most popular due to its compactness and efficiency. The skeleton, represented by a set of joint locations connected via limbs, is used to depict human pose.  This focuses exclusively on the human action without irrelevant details such as human appearance and environments. While encouraging results have been reported in the literature, the approach still struggles or fails in some common simple scenarios, mainly because it misses crucial action-related information pertaining to joint depth, body contour, and human-object interaction, as demonstrated in~\cref{fig:skl_fail}. 

To address the issue, several approaches have been proposed to extend skeleton by integrating object information~\cite{xuSkeletonBasedMutuallyAssisted2023, wangMultiStreamInteractionNetworks2022a, hachiumaUnifiedKeypointBasedAction2023a}. Despite their improved performance, critical information for HAR such as joint depth and body contour has not been considered. Moreover, object detection demands large amounts of additional annotations beyond skeleton extraction, hampering their deployability.

In this paper, we propose a simple yet effective representation called \modname{} to augment skeleton for video-based HAR. 
\modname{} is composed of three distinct binary maps, each capturing different action-related information:
\begin{itemize}
    \item \textit{Scale map volume} to capture joint depth: The scale (or size) of a joint on the camera plane provides valuable information on its depth. We propose a map volume for this scale, thereby capturing the depth information. 
    \item \textit{Body map} to capture body contour: We use the body map given by the contour of the entire human subject to cover body parts missing in skeleton. 
    \item \textit{Flow map} to capture human-object interaction: The interaction between the human subject and object provides 
    crucial clues on the human action. We propose using a flow map based on the optical flow values to reflect such interaction.
\end{itemize}

By augmenting the skeleton with all the above action-related information given by these maps, \modname{} offers a rich representation to achieve effective HAR.

To extract \modname{} from video, we present \netname{}, a novel \textbf{S}egmentation \textbf{Net}work supervised by the  optical \textbf{F}low and \textbf{S}keleton without additional labeling overhead. 
In \netname{}, the predictions of the scale map volume and body map are trained using the skeleton, while the optical flow predicted by a flow estimator supervises the flow map. 
By employing an unsupervised flow estimator, \netname{} requires no extra annotations for training beyond the skeleton approach, hence attaining training efficiency and facilitating wide deployment of \modname{}-based HAR pipeline. 

Our contributions are summarized as follows:
\begin{itemize}
    \item {\it \modname{}: A novel representation to augment skeleton for HAR:} 
    We propose a novel representation, \modname{}, to augment 2D skeleton for video-based HAR. With its rich action-related information, \modname{} combined with skeleton achieves higher HAR accuracy with similar compactness and efficiency as previous skeleton-only approaches.
    
    \item {\it \netname{}: An effective network to extract \modname{}:} We introduce \netname{}, which effectively predicts \modname{} from video utilizing only the point annotations from skeleton and optical flow without any additional annotation overhead. 
    
    \item {\it Extensive experiments and validation:} To validate our proposed pipeline based on \modname{} and its extractor \netname{}, we conduct extensive experiments on several commonly used datasets, namely  NTU RGB+D~\cite{shahroudyNTURGBLarge2016}, NTU RGB+D 120~\cite{liuNTURGB1202020a}, UCF101~\cite{soomroUCF101Dataset1012012}, and HMDB51~\cite{kuehneHMDBLargeVideo2011}. 
    Our results show that  \modname{}-augmented method significantly outperforms state-of-the-art skeleton-only approaches with comparable efficiency. Specifically, it achieves a substantial 2.2\% improvement in the X-Sub setting and a 1.6\% increase in the X-Set setting on NTU RGB+D 120.
\end{itemize}

\section{Related Works on Video-based HAR}
\label{sec:related}

\subsection{End-to-end Approaches}
The most straightforward HAR approach is to directly classify each input video into a specific human action category. Early works adopt spatiotemporal CNNs~\cite{tranCloserLookSpatiotemporal2018} or 3D-CNNs~\cite{carreiraQuoVadisAction2017, feichtenhoferSlowFastNetworksVideo2019, feichtenhoferX3DExpandingArchitectures2020a} as backbone, while transformer-based structure has become increasingly popular in recent years due to its ability to model long-term spatiotemporal dependency~\cite{arnabViViTVideoVision2021, piergiovanniRethinkingVideoViTs2023a, wangVideoMAEV2Scaling2023, srivastavaOmniVecLearningRobust2024, srivastavaOmniVec2NovelTransformer2024}. Although current video-based HAR methods achieve state-of-the-art performance on various datasets, their large amount of parameters, computational overhead and need for large amounts of data hamper their deployability in resource-constrained environment. 

\subsection{Skeleton-based HAR}
Compared with end-to-end methods, approaches using skeleton as an intermediate representation are considerably more compact and efficient, since the input size is reduced and noise from background and human appearance is eliminated. Recent years have witnessed the rapid evolution of skeleton-based HAR methods\cite{yanSpatialTemporalGraph2018a, choutasPoTionPoseMoTion2018, yanPA3DPoseAction3D2019, songConstructingStrongerFaster2023, wangNeuralKoopmanPooling2023a, Zhou_2025_ICCV, Liu_2025_CVPR}. Notably, PoseConv3D~\cite{duanRevisitingSkeletonbasedAction2022a} transforms skeletons into heatmaps and employs 3D-CNNs for prediction, achieving state-of-the-art accuracy and efficiency across multiple datasets. As part of the video-based HAR pipeline, these methods typically rely on 2D skeletons extracted from videos as input.
However, they suffer from significantly lower performance compared to end-to-end approaches in challenging cases, primarily due to the absence of crucial action-related information such as joint depth, body contour and human-object interaction. 

Some approaches incorporate object information into skeleton with additional annotations~\cite{xuSkeletonBasedMutuallyAssisted2023, wangMultiStreamInteractionNetworks2022a, hachiumaUnifiedKeypointBasedAction2023a}. Despite promising results, these methods require extra object detection, leading to significant additional annotation overhead and making them labor-intensive in practical applications. In contrast, our proposed \modname{} not only integrates more information, but also does not require annotations beyond skeleton extraction, enabling broader application scenarios.

\subsection{Optical Flow}
Optical flow is a dense vector field that describes motion on the camera plane between consecutive frames in a video. Flow estimators can be trained via unsupervised learning ~\cite{renUnsupervisedDeepLearning2017, stoneSMURFSelfTeachingMultiFrame2021, yuanUnSAMFlowUnsupervisedOptical2024}, requiring no annotation overhead. 

Early video-based HAR methods often leverage optical flow to extract motion information~\cite{simonyanTwoStreamConvolutionalNetworks2014a, crastoMARSMotionAugmentedRGB2019, sevilla-laraIntegrationOpticalFlow2019}. A more recent approach~\cite{Cai_2021_WACV} combines skeletons with joint-aligned optical flow patches to enhance understanding of human motion. Despite the improved performance, only optical flow around joints is used in this method, while valuable information from surrounding objects is not considered. Our \modname{}, in contrast, effectively utilizes optical flow to capture human-object interaction, thereby significantly improving the HAR performance.

\begin{figure}[t]
    \centering
    \includegraphics[width=\linewidth]{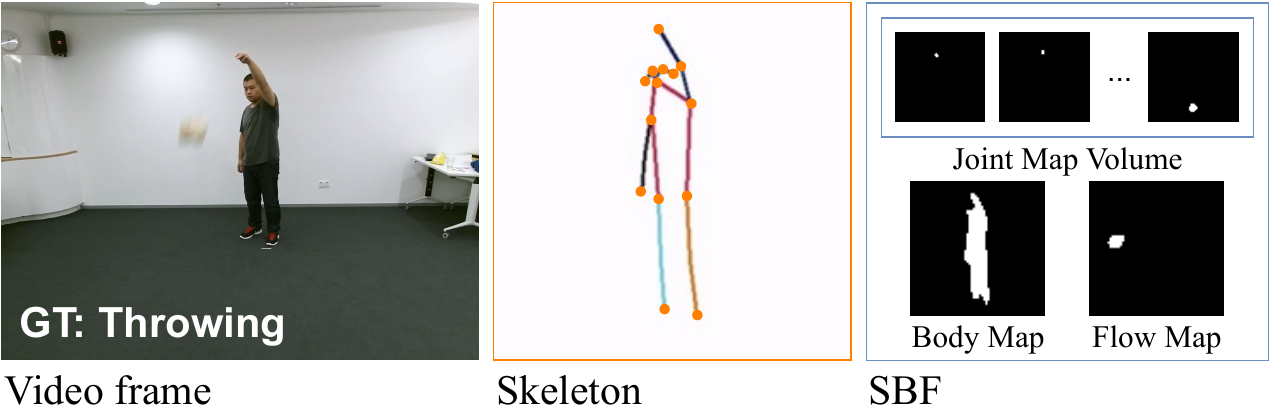}
    \caption{An example video frame, its extracted skeleton, and our proposed \modname{}.}
    \label{fig:sbf}
\end{figure}

\section{Scale-Body-Flow Representation}

We provide in this section a detailed description of Scale-Body-Flow (\modname{})
as a novel representation to augment 2D skeleton in the pipeline of video-based HAR. 
We begin with the definition of \modname{} in~\cref{sec:sbf_def}, as illustrated in~\cref{fig:sbf}. Next, we introduce the three \modname{} components, joint map volume (\cref{sec:sbf_joint}), body map (\cref{sec:sbf_body}) and flow map (\cref{sec:sbf_motion}). Finally, we explain how to combine \modname{} with skeleton for HAR and introduce our \modname{}-augmented HAR method, \schname{}, in~\cref{sec:sbf_gen}.


\subsection{Definition}\label{sec:sbf_def}
Given two consecutive video frames $\mathcal{I}_{t-1}$ and $\mathcal{I}_{t}$ of size $H_0\times W_0$, their \modname{} is composed of one scale map volume $\mathcal{S}_{t-1}\in\{0,1\}^{H\times W\times J}$, one body map $\mathcal{B}_{t-1}\in\{0,1\}^{H\times W}$ and one flow map $\mathcal{F}_t\in\{0,1\}^{H\times W}$, where $J$ is the number of joints, $H=H_0/4$ and $W=W_0/4$ empirically chosen to ensure compactness. We provide detailed description for these components in the following sections.

\subsection{Scale Map Volume}\label{sec:sbf_joint}
The scale map volume $\mathcal{S}$ consists of $J$ binary maps of size $H\times W$. $\mathcal{S}$ has two variants, the joint scale map volume $\mathcal{S}^J$ and the limb scale map volume $\mathcal{S}^L$. 

In $\mathcal{S}^J$, a pixel $x$ on the $i$-th map is assigned a value of 1 if $x$ belongs to the $i$-th joint in the skeleton, and 0 otherwise. The area of pixels with value 1 belonging to the joint, or ``scale'', is negatively correlated with joint depth, as objects closer to the camera appear larger on the image plane. 
Although predicting depth from video is challenging, $\mathcal{S}^J$ can be estimated using only joint locations, as we will elaborate in \cref{sec:snsnet_point}. 

$\mathcal{S}^L$ is derived from $\mathcal{S}^J$. Given the $i$-th limb connecting joints $a_i$ and $b_i$, the $i$-th map in $\mathcal{S}^L$ is defined as
\begin{align}
    \mathcal{S}_i^L(x)=\max(\mathcal{S}_{a_i}^J(x), \mathcal{S}_{b_i}^J(x)).
\end{align}

\subsection{Body Map}\label{sec:sbf_body}
Inspired by early works~\cite{CHAARAOUI20131799, uddin_human_2017}, we introduce body map $\mathcal{B}$, a binary map segmenting the human body from the background.
While its ground truth is not easily available, we will demonstrate how to approximate it using only skeleton during training in \cref{sec:snsnet_point}.

\begin{figure*}[t]
    \centering
    \includegraphics[width=0.75\linewidth]{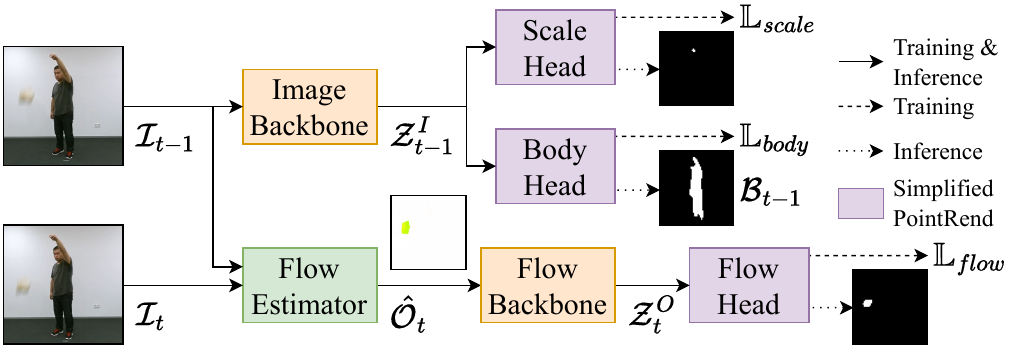}
    \caption{The overall structure of \netname{}. The flow estimator is pretrained via unsupervised learning.}
    \label{fig:snsnet}
\end{figure*}

\subsection{Flow Map}\label{sec:sbf_motion}
We observe from methods in unsupervised video object segmentation~\cite{luSeeMoreKnow2019, choTreatingMotionOption2023, leeUnsupervisedVideoObject2023} that pixels with optical flow distinct from the background often represent moving human body parts and interacting objects. Hence, human-object interaction can be inferred from optical flow values.

To capture this interaction, we introduce the flow map $\mathcal{F}$, a binary map derived from the optical flow $\mathcal{O}\in\mathbb{R}^{H\times W\times 2}$.
$\mathcal{F}$ is defined as follows:
\begin{align}\label{eq:flow_map}
   \Omega(x)&=\lVert \mathcal{O}(x) - \bar{\mathcal{O}} \rVert_{2},\\
   \mathcal{F}(x)&=
   \begin{cases}
    1,&\text{if }\Omega(x) > \epsilon \max{\Omega},\\
    0,&\text{otherwise.}
   \end{cases}
\end{align}
Here, $\bar{\mathcal{O}}$ is the average flow value representing the background, and $\epsilon>0$ is a constant that sets the threshold between static and moving objects. As we will discuss in~\cref{sec:snsnet_point}, optical flow can be predicted through unsupervised learning, allowing the estimation of $\mathcal{F}$ without requiring ground-truth optical flow $\mathcal{O}$.

\subsection{\schname{}: \modname{}-augmented HAR}\label{sec:sbf_gen}
Before combining \modname{} and skeleton, we first transform the skeleton into either a joint heatmap volume $\mathcal{H}^J$ or limb heatmap volume $\mathcal{H}^L$, following PoseConv3D~\cite{duanRevisitingSkeletonbasedAction2022a}. Both $\mathcal{H}^J$ and $\mathcal{H}^L$ are heatmaps of size $H\times W\times J$, generated using a 2D Gaussian filter with standard deviation $\sigma$. 
Using $\mathcal{H}^J$ and $\mathcal{S}^J$ results in the ``joint'' variant of our method, while $\mathcal{H}^L$ and $\mathcal{S}^L$ produce the ``limb'' variant. 
In the following, we use $\mathcal{H}$ and $\mathcal{S}$ to generically represent either the joint ($\mathcal{H}^J$, $\mathcal{S}^J$) or limb ($\mathcal{H}^L$, $\mathcal{S}^L$) variants.

The same Gaussian filter with standard deviation $\sigma$ is applied to the three components in \modname{} to obtain their smoothed versions: $\mathcal{S}'$, $\mathcal{B}'$ and $\mathcal{F}'$. The integrated volume $\mathcal{V}$ is calculated as a weighted sum of $\mathcal{H}$ and $\mathcal{S}'$:
\begin{align}
    \mathcal{V}=\mathcal{H}+\mu\mathcal{S}',
\end{align}
where $\mu$ is the weighting parameter. 
Finally, $\mathcal{V}$, $\mathcal{B}'$ and $\mathcal{F}'$ are concatenated to form a tensor of size $H\times W\times (J+2)$, which is fed into a downstream HAR model. Notably, we combine $\mathcal{H}$ and $\mathcal{S}'$ via a weighted sum because $\mathcal{H}_i$ and $\mathcal{S}'_j$ are highly correlated when $i=j$ but largely independent otherwise.  $\mathcal{V}$, $\mathcal{B}'$ and $\mathcal{F}'$ are concatenated because they provide complementary information for HAR.

Due to its similarity in size to $\mathcal{H}$, we employ the same 3D CNNs as PoseConv3D for our \modname{}-augmented HAR method, \textbf{\schname{}}, to validate the effectiveness of \modname{} independent of model architecture. Like previous multi-stream skeleton methods, \schname{} can integrate the ``joint'' and ``limb'' streams using both \modname{} variants.


\section{Extracting \modname{} with
\netname{} 
}\label{sec:snsnet}
In this section, we introduce \netname{}, an effective segmentation network for \modname{} prediction requiring no extra annotation overhead beyond the existing skeleton extraction. First,~\cref{sec:snsnet_over} presents a comprehensive overview of \netname{}. 
We then explain our Simplified PointRend (SPR) module for point-supervised segmentation and our proposed process for point annotation generation in~\cref{sec:snsnet_point}. Finally, the training details of \netname{} are provided in~\cref{sec:snsnet_train}.

\subsection{Overview}\label{sec:snsnet_over}
The overall structure of our \netname{} is illustrated in \cref{fig:snsnet}. 
The image backbone encodes an $\mathcal{I}_{t-1}$ into the image feature $\mathcal{Z}_{t-1}^I$, which is then fed into the scale head and body head to predict the joint scale volume $\mathcal{S}_{t-1}$ and the body map $\mathcal{B}_{t-1}$, respectively. Meanwhile, the optical flow $\mathcal{O}_{t}$ is predicted from $\mathcal{I}_{t-1}$ and $\mathcal{I}_{t}$ using an off-the-shelf flow estimator pretrained via unsupervised learning. The flow feature $\mathcal{Z}_{t}^O$ is then encoded from $\mathcal{O}_{t}$ with the flow backbone and subsequently input into the flow head to predict the flow map $\mathcal{F}_{t}$. The entire training process is supervised under three segmentation losses, namely $\mathbb{L}_{\mathit{scale}}$, $\mathbb{L}_{\mathit{body}}$ and $\mathbb{L}_{\mathit{flow}}$.

\subsection{Simplified PointRend (SPR)}\label{sec:snsnet_point}
We treat the prediction of each binary map of size $H\times W$ in \modname{} as a single-class segmentation task. 
To avoid the labor-intensive annotation process to obtain pixel-level ground truth for segmentation, we introduce our Simplified PointRend (SPR) module, which can utilize sparse point annotations for supervision instead of complete ground truths.
It employs a multi-layer perceptron that incorporates positional encoding and adaptive subdivision upsampling from Implicit PointRend (IPR)~\cite{kirillovPointRendImageSegmentation2020a, chengPointlySupervisedInstanceSegmentation2022a}, while excluding elements unnecessary for single-class segmentation. The scale head, body head
and flow head are three instances of the SPR module.

\begin{figure}[t]
    \centering
    \includegraphics[width=0.9\linewidth]{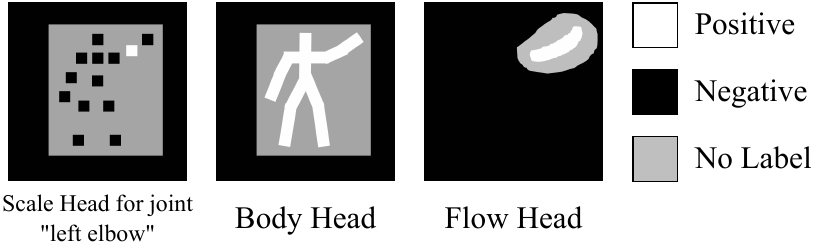}
    \caption{A conceptual example of the ``waving'' action for our annotation generation method in SPR. In each segmentation task, the white region denotes $\mathcal{P}^{\mathit{pos}}$ (positive labels), the black region represents $\mathcal{P}^{\mathit{neg}}$ (negative labels), and the grey region indicates areas excluded from point sampling.}
    \label{fig:pos_neg}
\end{figure}

All of our SPR modules are trained using point annotations derived from skeleton and optical flow, without requiring any additional annotation overhead beyond the existing skeleton extraction. 
The generation process for each SPR module begins by constructing positive and negative pixel sets, $\mathcal{P}^{\mathit{pos}}$ and $\mathcal{P}^{\mathit{neg}}$, with labels derived from skeleton or optical flow, as illustrated in~\cref{fig:pos_neg}. We then randomly sample a fixed number of positive and negative points from $\mathcal{P}^{\mathit{pos}}$ and $\mathcal{P}^{\mathit{neg}}$, respectively, as the point annotations used in training. 
Detailed explanations of the annotation generation process for the scale head, body head and flow head are provided below.

\begin{description}[style=unboxed, leftmargin=0cm]
\item[Scale Head:] 
We define the \textbf{joint set} of joint $i$, denoted as $\mathcal{P}_i^{joint}$, to be the set of points containing the $3\times 3$ pixels centered on its exact coordinate. The \textbf{background set} $\mathcal{P}^{\mathit{bg}}$ comprises all pixels outside the bounding box of the target individual with a padding of $\rho$. For each joint $i$, the positive set $\mathcal{P}_i^{\mathit{pos}}$ equals its joint set $\mathcal{P}_i^{joint}$, and the negative set is the union of all other joint sets and the background set:
\begin{align}
    \mathcal{P}_i^{\mathit{neg}} &= \left(\bigcup_{j\neq i}\mathcal{P}_j^{joint}\right)\cup\mathcal{P}^{\mathit{bg}}.
\end{align}
We sample $N^{\mathit{pos}}$ positive points from $\mathcal{P}_i^{\mathit{pos}}$ and $N^{\mathit{neg}}$ negative points from $\mathcal{P}_i^{\mathit{neg}}$. The training loss is calculated using the predicted binary classification scores $\mathcal{C}_{i}^{\mathit{pos}}$ at positive points and $\mathcal{C}_{i}^{\mathit{neg}}$ at negative points:
\begin{equation}
    \mathbb{L}_{\mathit{scale}} = \sum_{i}^{J} (BCE(\mathcal{C}_{i}^{\mathit{pos}}, \mathbf{I}) +\alpha\cdot BCE(\mathcal{C}_{i}^{\mathit{neg}}, \mathbf{O})),
\end{equation}
where $\alpha$ is a balancing constant, and $\mathbf{I}$ and $\mathbf{O}$ respectively denote the identity and zero tensor.

The $i$-th prediction of the scale head aggregates all pixels highly corresponding to the joint $i$ while excluding other body parts and the background. 
This ensures that the area of 1's in the $i$-th prediction roughly covers the entire joint $i$, thus estimating the $i$-th scale map $\mathcal{S}_{i}$ defined in~\cref{sec:sbf_joint}.

\item[Body Head:] We use Bresenham's line algorithm to connect joints on the image plane according to the graph structure of the skeleton. $\mathcal{P}^{\mathit{pos}}$ comprises all connected line segments and their adjacent pixels, while $\mathcal{P}^{\mathit{neg}}$ equals the background set as noted in the paragraph of Scale Head.
A constant of $N^{\mathit{body}}$ points are randomly sampled from either set. The loss function for the body head $\mathbb{L}_{\mathit{body}}$ is a BCE loss directly computed on all sampled points.

The body head prediction aims to distinguish pixels inside and outside the human body, with the boundary representing an approximate human body contour.

\item[Flow Head:] Points for the flow head are selected based on the predicted optical flow $\hat{\mathcal{O}}$ from an off-the-shelf unsupervised flow estimator. 
To generate $\mathcal{P}^{\mathit{pos}}$ and $\mathcal{P}^{\mathit{neg}}$, we first estimate $\Omega(x)$ from Eq. (2) by replacing the ground-truth $\mathcal{O}$ with its prediction $\hat{\mathcal{O}}$ and $\overline{\mathcal{O}}$ with $\overline{\hat{\mathcal{O}}}$:
\begin{align}
\hat{\Omega}(x) = \lVert\hat{\mathcal{O}}(x) - \bar{\hat{\mathcal{O}}}\rVert_{2},
\end{align}
where $\bar{\hat{\mathcal{O}}}$ is the mean flow vector of $\hat{\mathcal{O}}$. Given constants $\beta, \gamma\in (0, 1)$, 
$\mathcal{P}^{\mathit{pos}}$ and $\mathcal{P}^{\mathit{neg}}$ are defined as follows:
\begin{align}
    \mathcal{P}^{\mathit{pos}}&=\{x:\hat{\Omega}(x)>\beta\max\hat{\Omega}\},\\
    \mathcal{P}^{\mathit{neg}}&=\{x:\hat{\Omega}(x)<\gamma\max\hat{\Omega}\}.
\end{align}
$N^{\mathit{flow}}$ points are randomly sampled from either set and BCE is adopted as the loss function $\mathbb{L}_{\mathit{flow}}$.

The flow head prediction is clearly an estimate of $\mathcal{F}$ defined in~\cref{sec:sbf_motion}, with the threshold $\epsilon$ lying between $\beta$ and $\gamma$. 

\end{description}

\subsection{Model Training}\label{sec:snsnet_train}
We first train the flow backbone and flow head on a small video dataset with $\mathbb{L}_{\mathit{flow}}$ as the total loss. 
Then the remaining components of \netname{} are trained on a large-scale image dataset with the loss changed to
\begin{equation}
    \mathbb{L} = \mathbb{L}_{\mathit{scale}} + \lambda_{\mathit{body}}\mathbb{L}_{\mathit{body}},
\end{equation}
where $\lambda_{\mathit{body}}$ is the weighting parameter. 
\section{Experimental Evaluation}

\begin{table*}[t]
  \centering
  \small
  \caption{Comparison of model size, computational cost and accuracy (\%) with the state-of-the-art HAR methods on NTU and NTU120. All methods shown in the table extract intermediate representations (skeleton, object and \modname{}) from videos. Number of parameters and FLOPs are calculated for the downstream HAR model alone. Both 1-clip and 10-clip testing are applied to PoseConv3D and \schname{}. "J" and "L" indicate the joint and limb variants of SBF defined in~\cref{sec:sbf_gen}.}
  \begin{tabular}{c|ccccc|cccc}
    \toprule
    \multirow{2}{*}{Method} & \multirow{2}{*}{Category} & \multirow{2}{*}{Add. Anno.} & \multirow{2}{*}{Params (M)} & \multirow{2}{*}{FLOPs (G)} & \multirow{2}{*}{Clips} & \multicolumn{2}{c}{NTU} & \multicolumn{2}{c}{NTU120} \\
    & & & && & X-Sub & X-View & X-Sub & X-Set\\
    \midrule
    ST-GCN~\cite{yanSpatialTemporalGraph2018a} & Skeleton & \ding{55} & 12.3 & 15.3 & 10  & 92.4 & 98.3 & 84.7 & 89.0 \\
    AA-GCN~\cite{shiSkeletonBasedActionRecognition2020} & Skeleton & \ding{55} & 15.0 & 17.4 & 10  & 93.0 & 98.2 & 85.5 & 89.9 \\
    MS-G3D~\cite{liuDisentanglingUnifyingGraph2020a} & Skeleton & \ding{55} & 11.8 & 27.4 & 10  & 94.1 & 98.3 & 87.4 & 90.9 \\
    CTR-GCN~\cite{chenChannelwiseTopologyRefinement2021a} & Skeleton & \ding{55} & 5.7  & 7.8  & 10  & 93.6 & 98.4 & 86.6 & 90.1 \\
    ST-GCN++~\cite{duanPYSKLGoodPractices2022} & Skeleton & \ding{55} & 5.6  & 11.2 & 10  & 93.2 & 98.5 & 86.4 & 90.3 \\
    BlockGCN~\cite{zhouBlockGCNRedefineTopology2024} & Skeleton & \ding{55} & 5.2  & 8.4  & 10  & 93.9 & 98.2 & 87.3 & 90.7 \\
    ProtoGCN~\cite{Liu_2025_CVPR} & Skeleton & \ding{55} & 24.9 & 36.9 & 10  & 94.1 & \textbf{98.8} & 87.5 & 90.9 \\
    \midrule
    IOL~\cite{xuSkeletonBasedMutuallyAssisted2023} & Skl+Obj  & \ding{51} & -    & -    & -  & 90.0 & 95.7 & -    & -    \\
    MSI~\cite{wangMultiStreamInteractionNetworks2022a} & Skl+Obj  & \ding{51} & -    & -    & -  & 91.5 & 96.5 & 88.2 & 89.4 \\
    \midrule
    \multirow{2}{*}{PoseConv3D~\cite{duanRevisitingSkeletonbasedAction2022a}} & \multirow{2}{*}{Skeleton} & \multirow{2}{*}{\ding{55}} & \multirow{2}{*}{4.0} & \multirow{2}{*}{31.6} & 1  & 94.0 & 96.6 & 86.3 & 89.9 \\
                    & & & & & 10 & 94.1 & 97.1 & 86.9 & 90.3 \\
    \midrule
    \multirow{4}{*}{\schname{} (Ours)} & \multirow{2}{*}{Skl+\modname{}(J)} & \multirow{2}{*}{\ding{55}} & \multirow{2}{*}{2.0} & \multirow{2}{*}{16.2} & 1  & 94.6 & 97.8 & 89.0 & 91.5 \\
      & & & & & 10 & 94.6 & 98.0 & 89.1 & 91.7 \\
      & \multirow{2}{*}{Skl+\modname{}(J+L)} & \multirow{2}{*}{\ding{55}} & \multirow{2}{*}{4.0} & \multirow{2}{*}{32.4} & 1  & 94.8 & 97.9 & 89.4 & 92.2 \\
      & & & & & 10 & \textbf{95.0} & 98.1 & \textbf{89.6} & \textbf{92.3} \\
  \bottomrule
  \end{tabular}
  \label{tab:exp_acc}
\end{table*}

\subsection{Datasets}

We evaluate our method on four mainstream action recognition datasets: NTU RGB+D~\cite{shahroudyNTURGBLarge2016}, NTU RGB+D 120~\cite{liuNTURGB1202020a}, UCF101~\cite{soomroUCF101Dataset1012012} and HMDB51~\cite{kuehneHMDBLargeVideo2011}.

\begin{description}[style=unboxed, leftmargin=0cm]
\item[NTU RGB+D] (NTU) is a large-scale action recognition dataset collected in lab environment, with 56,880 action sequences performed by 40 subjects categorized into 60 classes. 
There are two standard benchmarks for this dataset, Cross-Subject (X-Sub) and Cross-View (X-View).

\item[NTU RGB+D 120] (NTU120) is an extended version of NTU RGB+D with 113,945 action sequences performed by 106 subjects categorized into 120 classes. There are two standard evaluation protocols most commonly used, namely Cross-Subject (X-Sub) and Cross-Setup (X-Set).

\item[UCF101 and HMDB51] are two video-based HAR datasets sourced from the Internet. UCF101 (UCF) contains 13,320 videos across 101 action categories, and HMDB51 (HMDB) includes 6,849 videos across 51 actions. 
In our evaluation, models trained on these datasets are pretrained on Kinetics 400~\cite{kayKineticsHumanAction2017}, a large-scale video action dataset. 

\end{description}

\subsection{Implementation Details}
\begin{description}[style=unboxed, leftmargin=0cm]
\item[\netname{}:] We use HRNet-W32~\cite{wangDeepHighResolutionRepresentation2021} as the image backbone, LiteHRNet-18~\cite{yuLiteHRNetLightweightHighResolution2021} as the flow backbone, and SMURF~\cite{stoneSMURFSelfTeachingMultiFrame2021} is selected as the flow estimator. 
The data augmentation pipeline follows~\cite{kirillovPointRendImageSegmentation2020a}
, with each bounding box padded to a square and resized to $256\times 256$. During the process of point annotation generation, we use padding $\rho=10$, $N_{pos}=32, N_{neg}=128, N_{body}=N_{flow}=256$, $\alpha=19$, $\beta=0.8$, and $\gamma=0.2$. The loss weights are set to $\lambda_{joint}=0.025$ and $\lambda_{body}=1$. 
We first train the flow backbone and flow head in \netname{} on J-HMDB~\cite{jhuangUnderstandingActionRecognition2013} for 100 epochs and then train the other components on COCO17~\cite{linMicrosoftCOCOCommon2014} for 210 epochs with the flow backbone and flow head frozen.

\item[\schname{}:] We employ SlowOnly-R50 (SO-R50)~\cite{feichtenhoferSlowFastNetworksVideo2019} as the 3D CNN backbone and adhere to the training pipeline in~\cite{duanRevisitingSkeletonbasedAction2022a} for our \modname{}-augmented HAR. 
The length of each \modname{} sequence is 48, and the crop size is $56\times 56$ during training. 
$\mu$ is set to 0.1, and a Gaussian filter with $\sigma=0.4$ is applied. 
The models are trained for 240 epochs in total. 
In our experiments, we apply 10-clip testing, which aggregates results of 10 distinct samples from a video, unless otherwise specified.
When fusing the two streams of \schname{}, we average their prediction scores to produce the final results.
\end{description}

\subsection{Comparison with State of the Art}
In this section, we compare our \modname{}-augmented HAR (Skl+\modname{}) with state-of-the-art HAR approaches based on intermediate representations extracted from videos. 
Selected baselines belong to two categories: 
\begin{description}[style=unboxed, leftmargin=0cm]
\item[Skeleton methods], including ST-GCN~\cite{yanSpatialTemporalGraph2018a}, AA-GCN~\cite{shiSkeletonBasedActionRecognition2020}, MS-G3D~\cite{liuDisentanglingUnifyingGraph2020a}, CTR-GCN~\cite{chenChannelwiseTopologyRefinement2021a}, ST-GCN++~\cite{duanPYSKLGoodPractices2022}, BlockGCN~\cite{zhouBlockGCNRedefineTopology2024}, PoTion~\cite{choutasPoTionPoseMoTion2018}, PA3D~\cite{yanPA3DPoseAction3D2019} and PoseConv3D~\cite{duanRevisitingSkeletonbasedAction2022a}.
For a fair comparison, we evaluate their performance using 2D skeleton extracted from videos using HRNet-w32~\cite{wangDeepHighResolutionRepresentation2021} with a clip length of 100. BlockGCN is trained by ourselves, while the results for other such methods are sourced from~\cite{duanPYSKLGoodPractices2022}. For methods involving multi-stream fusion, we report results using the most streams.
\item[Skeleton-object fusion methods]~(Skl+Obj) which integrate object information into 2D skeleton using additional annotations, including IOL~\cite{xuSkeletonBasedMutuallyAssisted2023}, MSI~\cite{wangMultiStreamInteractionNetworks2022a}, and SKP~\cite{hachiumaUnifiedKeypointBasedAction2023a}. We present the best results reported in their papers. 
\end{description}

\begin{table*}[t]
\begin{minipage}{.4\textwidth}
  \small
  \centering
  \caption{Comparison of accuracy (\%) with the state-of-the-art HAR methods on UCF and HMDB. }
  \resizebox{\linewidth}{!}{
  \begin{tabular}{c|c|cc}
    \toprule
    Method          & Category   & UCF  & HMDB \\
    \midrule
    PoTion~\cite{choutasPoTionPoseMoTion2018} & Skeleton   & 65.2 & 43.7 \\
    PA3D~\cite{yanPA3DPoseAction3D2019} & Skeleton   & -    & 55.3 \\
    \midrule
    SKP~\cite{hachiumaUnifiedKeypointBasedAction2023a} & Skl+Obj    & \textbf{87.8} & 70.9 \\
    \midrule
    PoseConv3D~\cite{duanRevisitingSkeletonbasedAction2022a} & Skeleton   & 87.0 & 69.7 \\
    \midrule
    \schname{}      & Skl+\modname{} & 87.5 & \textbf{71.2} \\
  \bottomrule
  \end{tabular}}
  \label{tab:exp_acc_k400}
\end{minipage}
\hspace{0.05\textwidth}
\begin{minipage}{.55\textwidth}
  \small
  \centering
  \caption{Comparison of efficiency of entire video-based HAR pipelines on NTU X-Sub. Number of parameters and FLOPs are calculated for both the skeleton/\modname{} extraction network and the downstream HAR model. 1-clip testing is used here. E. V. stands for ``extractor variant''.
  }
  \resizebox{\linewidth}{!}{
  \begin{tabular}{c|cccc|c}
    \toprule
    Method          & Category   & E. V. & Params & FLOPs & Acc (\%) \\
    \midrule
    CTR-GCN~\cite{chenChannelwiseTopologyRefinement2021a} & Skeleton   & Base & 36.3M       & 1.1T       & 93.5 \\    
    Proto-GCN~\cite{Liu_2025_CVPR} & Skeleton   & Large   & 90.6M       & 3.6T       & 94.2 \\    
    \midrule
    \multirow{2}{*}{\schname{}} & \multirow{2}{*}{Skl+\modname{}} & Small & \textbf{25.7M} & \textbf{617G} & 93.8 \\
     & & Base  & 67.4M       & 3.5T       & \textbf{94.8} \\
  \bottomrule
  \end{tabular}}
  \label{tab:exp_effi}
\end{minipage}
\end{table*}

As shown in~\cref{tab:exp_acc}, our \schname{} with 1-clip testing already outperforms all other methods in 3 out of 4 settings on NTU and NTU120. Notably, \schname{}'s performance even exceeds skeleton-object fusion methods, which rely on additional annotations, across all settings. With 10-clip testing, \schname{} achieves a 2.2\% 
accuracy increase on NTU120 X-Sub and a 1.6\% increase on NTU120 X-Set compared to skeleton methods. Results in~\cref{tab:exp_acc_k400} also demonstrate that our \schname{} attains higher performance than skeleton methods on UCF and HMDB and even surpasses SKP on HMDB, indicating its effectiveness in more challenging scenarios. Even in the NTU X-view setting, where \schname{} shows lower accuracy than state-of-the-art skeleton approaches, it still substantially outperforms PoseConv3D, which utilizes the same model structure and training pipeline, underscoring that \modname{} effectively complements the skeleton with rich action representation. 

Beyond performance, we assess the efficiency of \netname{} and \schname{} in comparison with state-of-the-art skeleton approaches. 
\cref{tab:exp_acc} presents the statistics for the downstream HAR model alone. The results reveal that \schname{} achieves the highest accuracy with the smallest model size and comparable computational cost in most settings on NTU and NTU120. 
\cref{tab:exp_effi} further exhibits the parameter count and FLOPs for the entire video-based HAR pipeline, including both the skeleton/\modname{} extraction and the downstream HAR model, with 1-clip testing. 
In addition to the Base extractor variants used in~\cref{tab:exp_acc}, we employ a larger skeleton extractor for ProtoGCN (Large) and a more compact skeleton+\modname{} extractor for \schname{} (Small) to further highlight the efficiency of \modname{}. The former adopts HRNet-w48~\cite{wangDeepHighResolutionRepresentation2021} with 4x higher input resolution; the latter uses RTM-Pose~\cite{jiangRTMPoseRealTimeMultiPerson2023a} for skeleton extraction, LiteHRNet-30~\cite{yuLiteHRNetLightweightHighResolution2021} as the image backbone, LiteHRNet-18~\cite{yuLiteHRNetLightweightHighResolution2021} as the flow backbone, and downsamples the input size to $1/16$ in the flow estimator. 
Results show that Small \schname{} achieves higher accuracy than Base CTR-GCN with lower computational cost, while Base \schname{} outperforms Large ProtoGCN with comparable efficiency.

\begin{table}[t]
  \centering
  \caption{Comparison of accuracy (\%) on videos captured from different viewing angles on NTU120 X-Sub. }
  \begin{tabular}{c|c|ccc}
    \toprule
    Method          & Category   & Front & Side & $45^{\circ}$ \\
    \midrule
    PoseConv3D~\cite{duanRevisitingSkeletonbasedAction2022a} & Skeleton   & 87.4  & 83.5 & 86.9 \\
    \schname{} & Skl+\modname{}& \textbf{90.1}  & \textbf{87.3} & \textbf{89.9} \\
    $\Delta$        & -          & +2.7  & +3.8 & +3.0 \\
  \bottomrule
  \end{tabular}
  \label{tab:exp_view}
\end{table}

\begin{table}[t]
  \centering
  \caption{Comparison of accuracy (\%) on actions with different difficulty levels on NTU120 X-Sub. }
  \resizebox{\linewidth}{!}{
  \begin{tabular}{c|c|ccc}
    \toprule
    Method          & Category   & Hard & Medium & Easy \\
    \midrule
    PoseConv3D~\cite{duanRevisitingSkeletonbasedAction2022a} & Skeleton   & 52.2 & 82.2   & 96.5 \\
    \schname{} & Skl+\modname{} & \textbf{61.1} & \textbf{87.1}   & \textbf{97.4} \\
    $\Delta$        & -          & +8.9 & +4.9   & +0.9 \\
  \bottomrule
  \end{tabular}}
  \label{tab:exp_hard}
\end{table}

\begin{figure}[t]
    \centering
    \includegraphics[width=\linewidth]{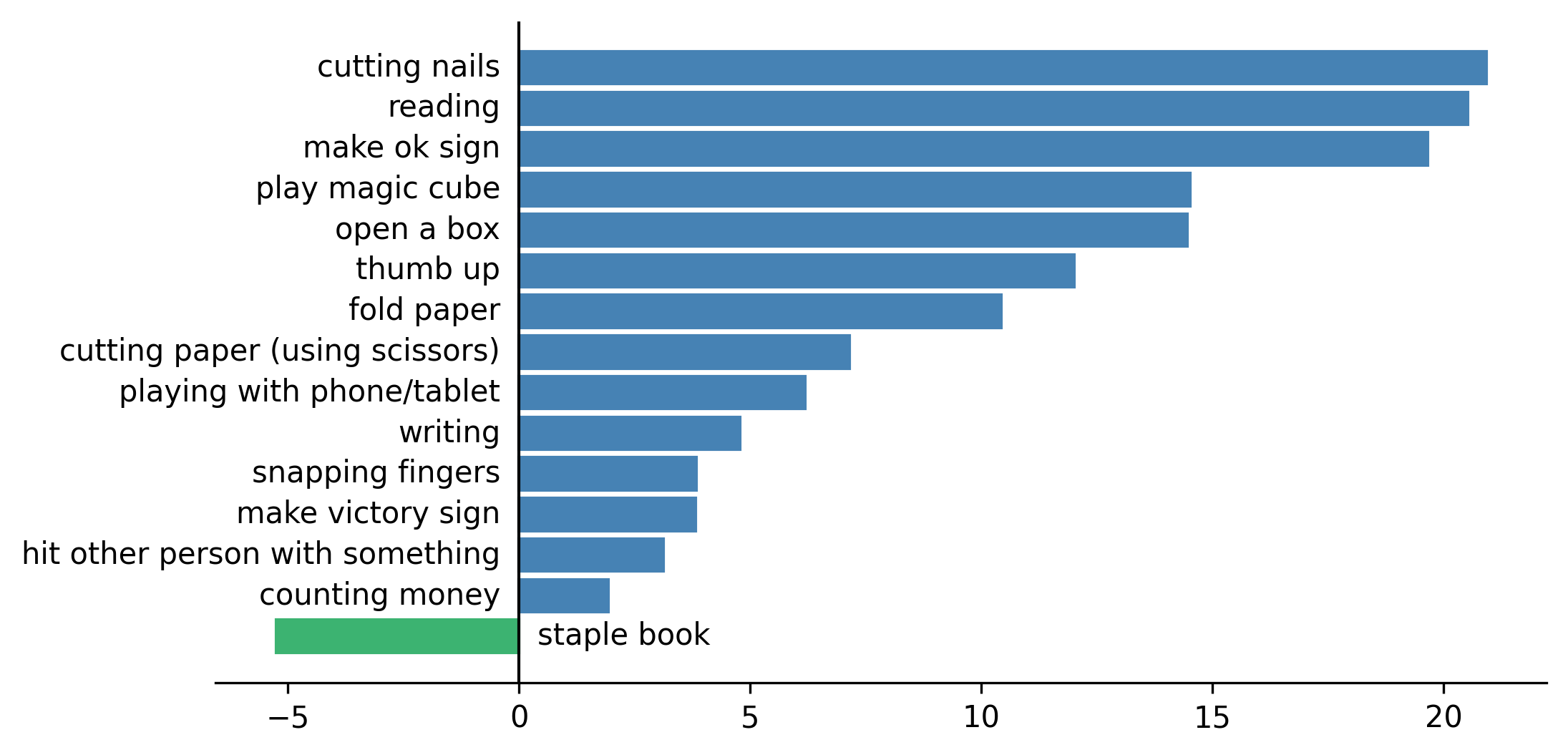}
    \caption{The accuracy difference (\%) between our \schname{} and PoseConv3D for hard actions on NTU120 X-Sub.}
    \label{fig:hard_actions}
\end{figure}

\subsection{Study on 
Challenging Scenarios}\label{sec:challenging}
To more clearly demonstrate the advantage of \modname{}-augmented HAR, we conduct further analysis on the recognition results in challenging scenarios characterized by difficult viewing angles or hard action categories. 
Only the joint stream is used for \schname{} and PoseConv3D.

\begin{description}[style=unboxed, leftmargin=0cm]
\item[Viewing angles:] We divide the testing data in NTU120 X-Sub into three splits according to their viewing angles, namely the Front, Side and $45^{\circ}$ view. Intuitively, data from the Side view suffers the most from lack of depth information, as it provides the least amount of spatial differentiation. 
Results in~\cref{tab:exp_view} confirm this intuition, showing lowest accuracy for the Side view. Our \schname{} attains greater performance improvement on the Side view (+3.8\%) than on Front (+2.7\%) and $45^{\circ}$ (+3.0\%) views, which demonstrates that the joint scale map volume effectively provides depth information to enhance the action representation.
\item[Action Categories:] Following~\cite{zhouLearningDiscriminativeRepresentations2023a}, we classify the action categories in NTU120 into three difficulty levels indicated by the action-specific accuracy of PoseConv3D. Actions with accuracy above 90\% are considered as Easy, those within 70-90\% as Medium, and those below 70\% as Hard. 
As demonstrated in \cref{tab:exp_hard}, our \schname{} significantly boosts the accuracy for  Hard (+8.9\%) and Medium (+4.9\%) actions. \cref{fig:hard_actions} lists all the Hard actions and shows accuracy improvements in 14 out of the 15 actions, including those highly related to body contour (\eg ``make OK sign'') or human-object interaction (\eg ``cutting nails''). This suggests that the body map and flow map in \modname{} effectively capture the relevant information, substantially enhancing HAR performance in challenging scenarios.
\end{description}

\subsection{Ablation Studies and Qualitative Analysis
}
This section presents ablation studies and qualitative analysis using the joint stream of \schname{} and PoseConv3D.

\begin{description}[style=unboxed, leftmargin=0cm]
\item[Effects of \modname{} Components:] We validate the effectiveness of all three components of \modname{}, scale map volume $\mathcal{S}$, body map $\mathcal{B}$ and flow map $\mathcal{F}$ on NTU X-Sub. Results in~\cref{tab:abl_comp} demonstrate that all components collectively enhance the performance of our \modname{}-augmented HAR. Notably, the flow map contributes the most, possibly because NTU includes various types of objects.

\item[Comparison with PoseConv3D with different 3D CNNs:] In addition to SO-R50, we also compare the performance of our \schname{} with PoseConv3D using two other 3D CNNs, namely C3D-s~\cite{tranLearningSpatiotemporalFeatures2015} and X3D-s~\cite{feichtenhoferX3DExpandingArchitectures2020a}, on NTU X-Sub. As shown in~\cref{tab:abl_model}, \schname{} outperforms PoseConv3D across all three model types, with only a marginal increase in model size and computational cost. This compelling evidence proves that \modname{} improves HAR accuracy with comparable efficiency regardless of the model structure.

\begin{table}
      \centering
      \caption{Ablation Study on \modname{} components on NTU X-Sub. }
      \begin{tabular}{cccc|c}
        \toprule
        Skeleton & $\mathcal{S}$ & $\mathcal{B}$ & $\mathcal{F}$ & Acc (\%)\\
        \midrule
        \ding{51} &           &           &           & 93.5 \\
                  & \ding{51} &           &           & 90.3 \\
                  &           & \ding{51} &           & 86.7 \\
                  &           &           & \ding{51} & 87.8 \\
        \midrule
        \ding{51} & \ding{51} &           &           & 93.6 \\
        \ding{51} & \ding{51} & \ding{51} &           & 93.8 \\
        \ding{51} & \ding{51} &           & \ding{51} & 94.5 \\
        \midrule
        \ding{51} & \ding{51} & \ding{51} & \ding{51} & \textbf{94.6} \\    
      \bottomrule
      \end{tabular}
      \label{tab:abl_comp}
\end{table}

\begin{table}
    \small
    \centering
    \caption{Comparison with PoseConv3D using different downstream networks on NTU X-Sub. }
    \resizebox{\linewidth}{!}{
    \begin{tabular}{c|c|ccc}
      \toprule
      Method                      & Network & Acc (\%) & Params & FLOPs\\
      \midrule
      \multirow{3}{*}{PoseConv3D~\cite{duanRevisitingSkeletonbasedAction2022a}} & SO-R50  & 93.7 & 2.0M  & 15.8G \\
                                  & C3D-s   & 92.9 & 3.4M  & 16.8G \\
                                  & X3D-s   & 92.3 & 241K  & 0.6G  \\
      \midrule
      \multirow{3}{*}{\schname{}} & SO-R50  & \textbf{94.6} & 2.0M  & 16.2G \\
                                  & C3D-s   & \textbf{93.4} & 3.4M  & 17.1G \\
                                  & X3D-s   & \textbf{92.4} & 242K  & 0.7G  \\
    \bottomrule
    \end{tabular}}
    \label{tab:abl_model}
  \end{table}

\begin{figure}[t]
    \centering
    \begin{subfigure}{.23\linewidth}
        \centering
        \includegraphics[width=\linewidth]{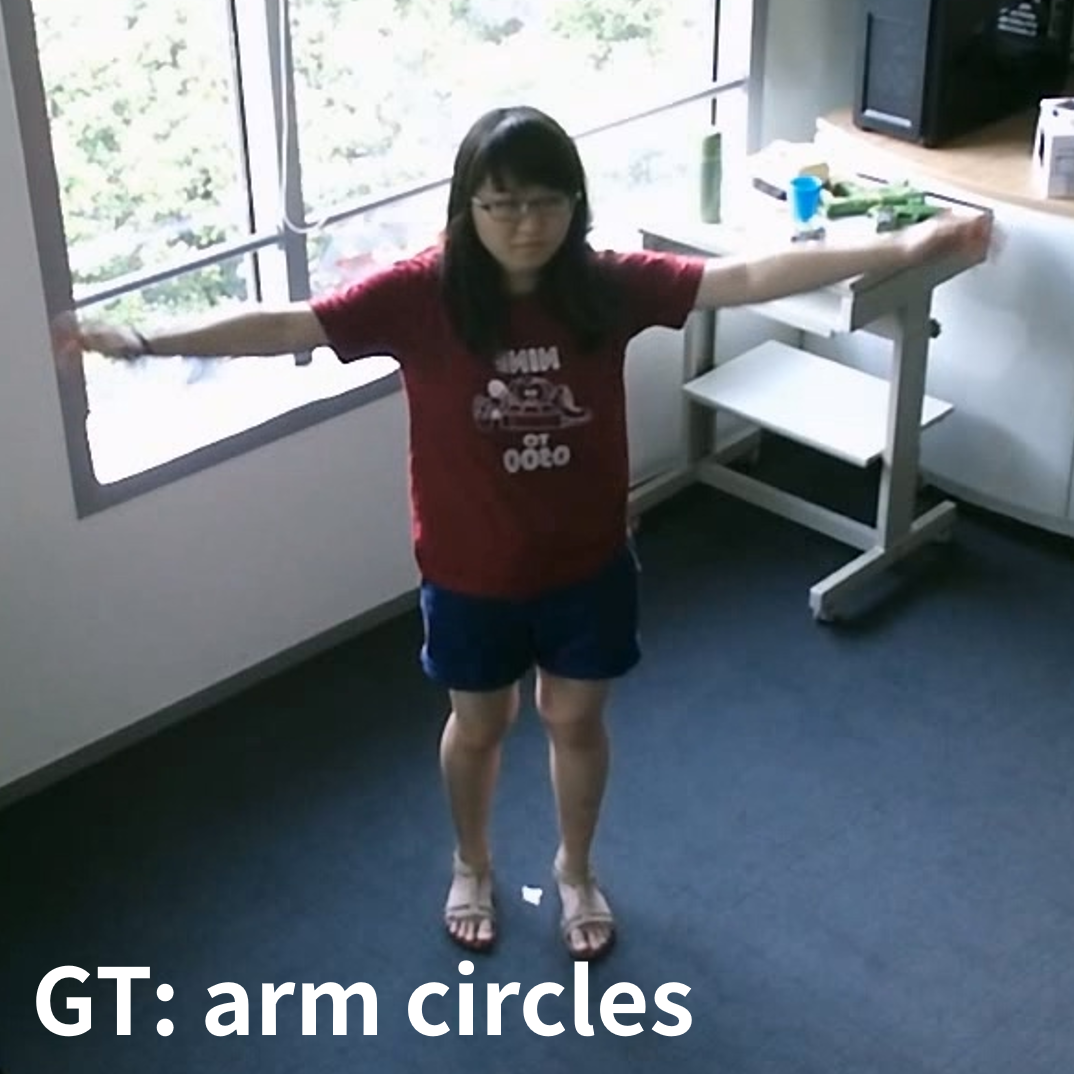}
    \end{subfigure}
    \begin{subfigure}{.23\linewidth}
        \centering
        \includegraphics[width=\linewidth]{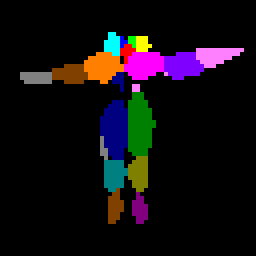}
    \end{subfigure}
    \begin{subfigure}{.23\linewidth}
        \centering
        \includegraphics[width=\linewidth]{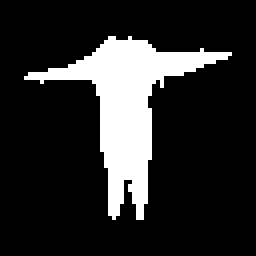}
    \end{subfigure}
    \begin{subfigure}{.23\linewidth}
        \centering
        \includegraphics[width=\linewidth]{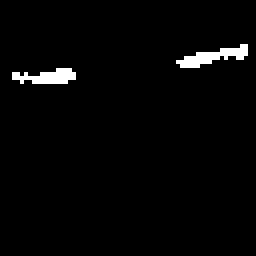}
    \end{subfigure}
    \begin{subfigure}{.23\linewidth}
        \centering
        \includegraphics[width=\linewidth]{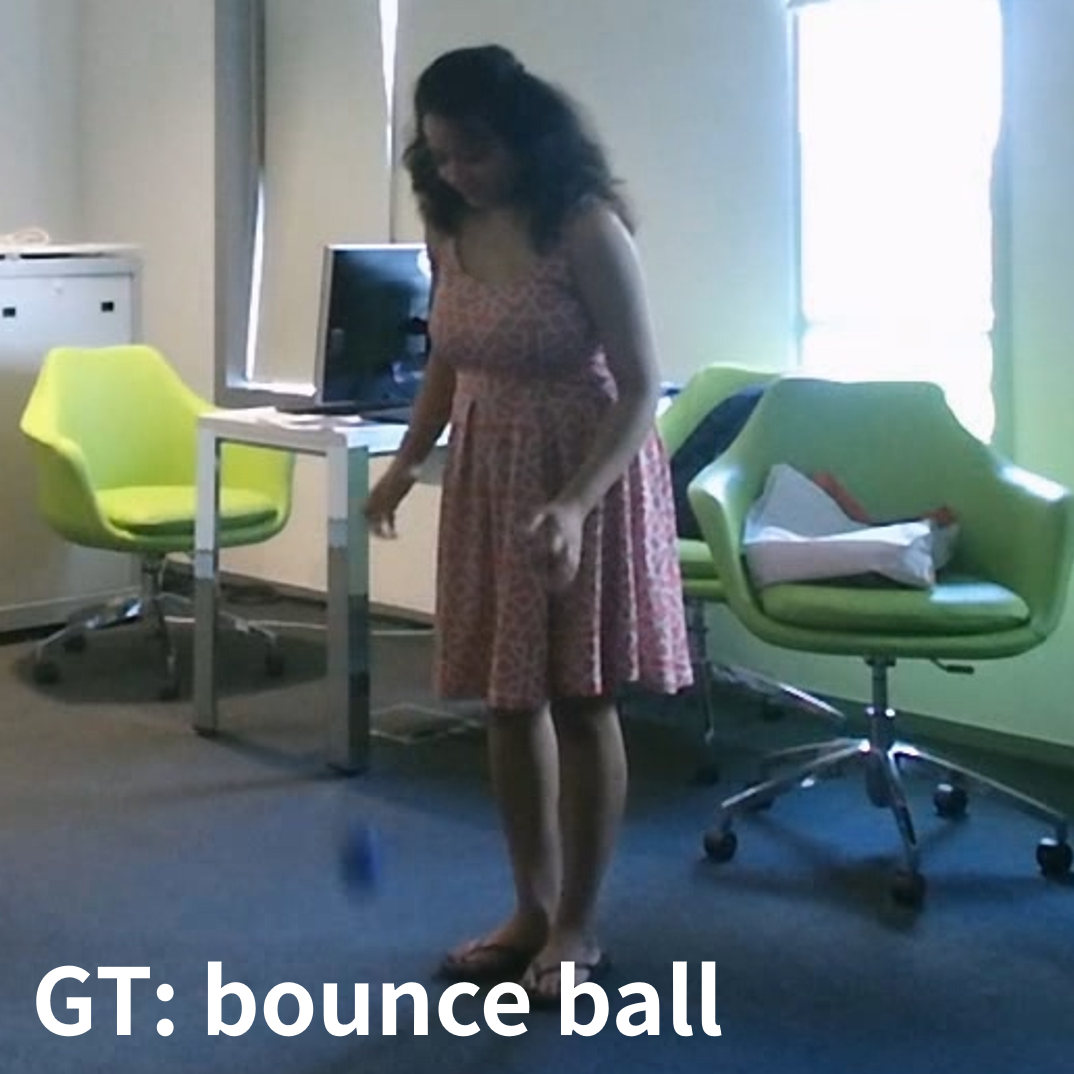}
    \end{subfigure}
    \begin{subfigure}{.23\linewidth}
        \centering
        \includegraphics[width=\linewidth]{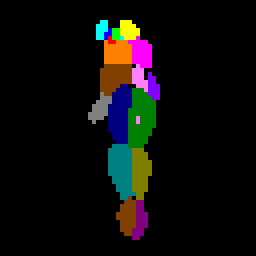}
    \end{subfigure}
    \begin{subfigure}{.23\linewidth}
        \centering
        \includegraphics[width=\linewidth]{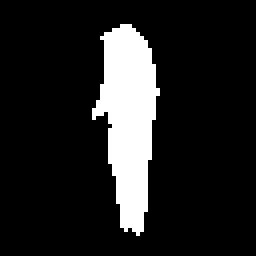}
    \end{subfigure}
    \begin{subfigure}{.23\linewidth}
        \centering
        \includegraphics[width=\linewidth]{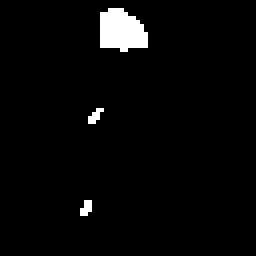}
    \end{subfigure}
    \begin{subfigure}[t]{.23\linewidth}
        \centering
        \includegraphics[width=\linewidth]{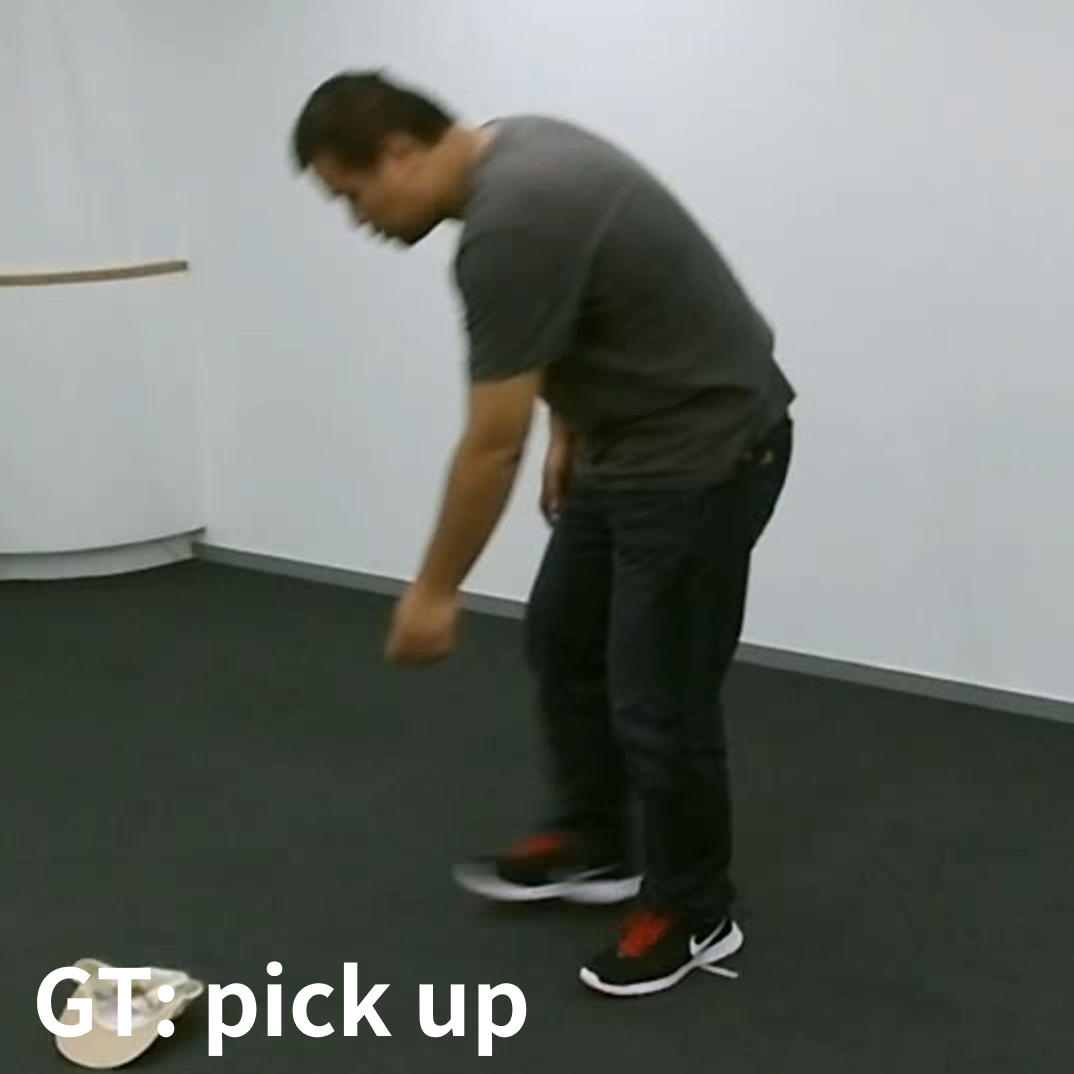}
        \caption{Image}
    \end{subfigure}
    \begin{subfigure}[t]{.23\linewidth}
        \centering
        \includegraphics[width=\linewidth]{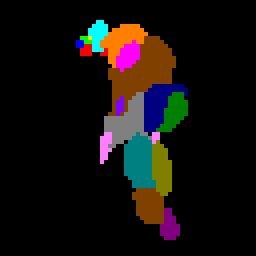}
        \caption{$\mathcal{S}$}
    \end{subfigure}
    \begin{subfigure}[t]{.23\linewidth}
        \centering
        \includegraphics[width=\linewidth]{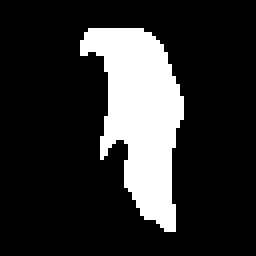}
        \caption{$\mathcal{B}$}
    \end{subfigure}
    \begin{subfigure}[t]{.23\linewidth}
        \centering
        \includegraphics[width=\linewidth]{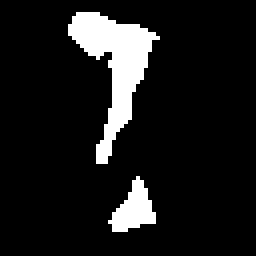}
        \caption{$\mathcal{F}$}
    \end{subfigure}

    \caption{Visualization of \modname{} components predicted by \netname{} on NTU120 X-Sub. Each joint's scale map has a distinct color.}
    \label{fig:vis}
\end{figure}

\item[Visualization of Predicted \modname{}:] We present visualization of \modname{} samples predicted by \netname{} in~\cref{fig:vis}. The results demonstrate that $\mathcal{S}$ effectively reflects the scale of each joint, $\mathcal{B}$ successfully encircles various body parts, and $\mathcal{F}$ clearly captures the moving body parts and interacting objects. This validates that \modname{} effectively incorporates the action-related information. 
\end{description}
\section{Conclusion}
Previous works on video-based human action recognition (HAR) are usually based on 2D skeleton as an intermediate representation, which struggles with common scenes due to its lack of action-related information, such as joint depth, body contour and human-object interaction.  
We propose a novel representation called \modname{} to augment skeleton
for video-based HAR. \modname{} consists of a scale map volume with the scale (relating to depth) of each joint, a body map outlining the human contour, and a flow map based on optical flow values to capture 
the interaction between the human and the object. We further present the extractor \netname{}, which predicts \modname{} with point annotations generated from skeleton and unsupervised optical flow. Without additional annotation overhead beyond existing skeleton extraction, \modname{} effectively integrates the action-related information with similar compactness and efficiency. 
Extensive experimental results on commonly used datasets show that our \modname{}-augmented HAR pipeline outperforms state-of-the-art skeleton methods with similar efficiency. 
Future work may explore more advanced designs for \netname{} to improve the quality of predicted \modname{}, and develop more powerful downstream models for \modname{}-augmented HAR.

\section{Acknowledgment}
This work was supported,in part, by RGC-General Research Fund (under grant number 16201625),
and Smart Traffic Fund (under grant number STF26EG01) of Hong Kong.
{
    \small
    \bibliographystyle{ieeenat_fullname}
    \bibliography{main}
}

\clearpage
\setcounter{page}{1}
\maketitlesupplementary

\begin{table*}[t]
  \begin{minipage}{.65\textwidth}
  \begin{tabular}{ccccccc}
    \toprule
    \multirow{2}{*}{Method} & \multirow{2}{*}{Category} & \multirow{2}{*}{Clips} & \multicolumn{2}{c}{NTU} & \multicolumn{2}{c}{NTU120} \\
    & & & X-Sub & X-View & X-Sub & X-Set\\
    \midrule
    \multirow{2}{*}{PoseConv3D} & \multirow{2}{*}{Skeleton (Limb)} & 1 & 93.2 & 95.7 & 85.7 & 89.4 \\
    & & 10 & 93.4 & 96.0 & 85.9 & 89.7 \\
    \midrule
    \multirow{2}{*}{\schname{}} 
      & \multirow{2}{*}{\modname{}} & 1 & 94.4 & 96.4 & 88.5 & 91.3 \\
      & & 10 & \textbf{94.4} & \textbf{96.6} & \textbf{88.6} & \textbf{91.6} \\
  \bottomrule
  \end{tabular}
  \caption{Comparison of accuracy (\%) with PoseConv3D using only the ``limb'' variant. }
  \label{tab:exp_lbf}
  \end{minipage}
  \begin{minipage}{.35\textwidth}
  \centering
  \begin{tabular}{cc}
    \toprule
    $\mu$ & Acc (\%)\\
    \midrule
    0     & 94.5 (.46) \\
    0.05  & 94.5 (.48)\\
    0.1   & \textbf{94.6} \\
    0.5   & 93.9 \\
    1     & 93.3 \\
  \bottomrule
  \end{tabular}
  \caption{Ablation Study on $\mu$. }
  \label{tab:abl_mu}
  \end{minipage}
\end{table*}

\begin{table*}[t]
\begin{minipage}{.47\textwidth}
  \centering
  \begin{tabular}{ccccc}
    \toprule
    Method & Category & Add. Anno. & Acc (\%) \\
    \midrule
    \multirow{2}{*}{CTR-GCN} & 2D Skl & \ding{55} & 93.6\\
                    & GT 3D Skl & \ding{51} & 92.1\\
                    & Pred 3D Skl & \ding{51} & 63.9\\
    \midrule
    \schname{} & \modname{} & \ding{55} & \textbf{95.0}\\
  \bottomrule
  \end{tabular}
  \caption{Comparison of accuracy (\%) with skeleton-based methods trained using 3D skeletons. }
  \label{tab:exp_3d}
\end{minipage}
\begin{minipage}{.52\textwidth}
  \centering
  \begin{tabular}{ccccc}
    \toprule
    Method & Category & Clip Len. & X-Sub & X-View \\
    \midrule
    CTR-GCN & Skeleton & 48 & 92.6 & 97.8\\
    PoseConv3D & Skeleton & 48 & 94.1 & 97.1\\
    \schname{} & \modname{} & 48 & \textbf{95.0} & \textbf{98.1} \\
  \bottomrule
  \end{tabular}
  \caption{Comparison of accuracy (\%) with skeleton-based methods with the same number of frames per clip. }
  \label{tab:exp_cl}
\end{minipage}
\end{table*}

\begin{table}[t]
  \centering
  \begin{tabular}{ccccc}
    \toprule
    Skeleton & $\mathcal{S}$ & $\mathcal{B}$ & $\mathcal{F}$ & Acc (\%) \\
    \midrule
    \ding{51} &           &           &           & 93.5 \\
              & \ding{51} &           &           & 90.3 \\
              &           & \ding{51} &           & 86.7 \\
              &           &           & \ding{51} & 87.8 \\
    \midrule
    \ding{51} & \ding{51} &           &           & 93.6 \\
    \ding{51} &           & \ding{51} &           & 93.6 \\
    \ding{51} &           &           & \ding{51} & 94.3 \\
              & \ding{51} & \ding{51} &           & 91.1 \\
              & \ding{51} &           & \ding{51} & 92.3 \\
              &           & \ding{51} & \ding{51} & 89.8 \\
    \midrule                                            
    \ding{51} & \ding{51} & \ding{51} &           & 93.8 \\
    \ding{51} & \ding{51} &           & \ding{51} & 94.5 \\
    \ding{51} &           & \ding{51} & \ding{51} & 94.5 \\
              & \ding{51} & \ding{51} & \ding{51} & 93.3 \\
    \midrule                                            
    \ding{51} & \ding{51} & \ding{51} & \ding{51} & \textbf{94.6} \\
  \bottomrule
  \end{tabular}
  \caption{Complete Ablation Study on \modname{} components on NTU X-Sub. }
  \label{tab:abl_comp2}
\end{table}

\section{More Implementation Details}
In this section, we elaborate more details of our implementations of \netname{} and \schname{}. All our experiments are conducted on two hardware platforms, one with 8 NVIDIA GeForce 2080Ti GPUs and 16 CPUs and the other with 4 3090Ti GPUs and 40 CPUs. 

\subsection{\netname{}}
\begin{figure}[h]
    \centering
    \begin{subfigure}{\linewidth}
        \centering
        \includegraphics[width=\linewidth]{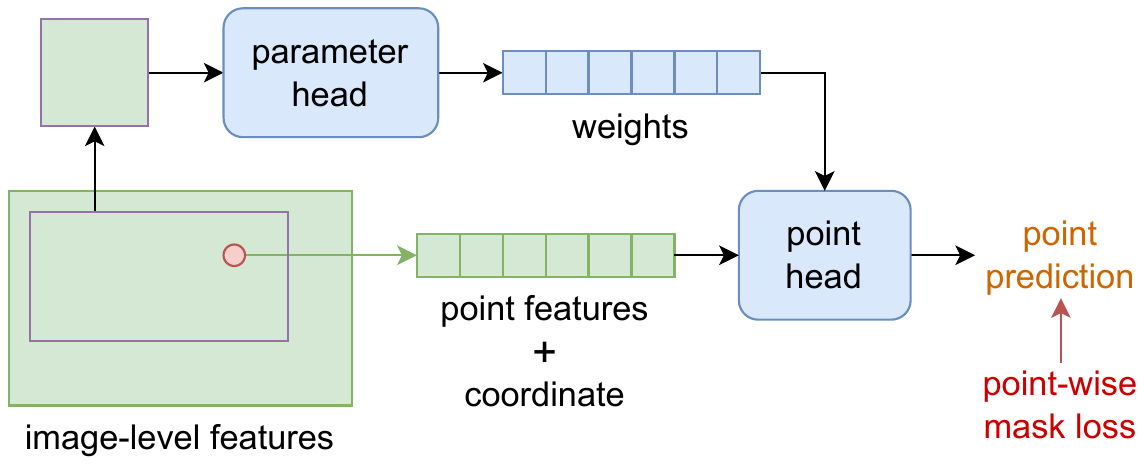}
        \caption{Implicit PointRend}
    \end{subfigure}
    \begin{subfigure}{\linewidth}
        \centering
        \includegraphics[width=\linewidth]{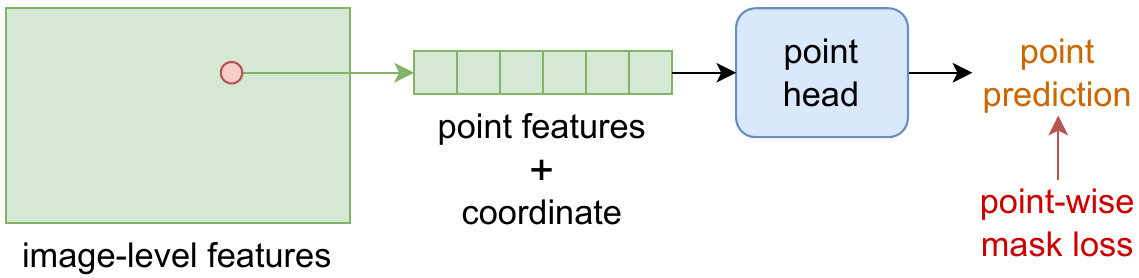}
        \caption{Simplified PointRend (Ours)}
    \end{subfigure}
    \caption{Comparison of Implicit PointRend~\cite{chengPointlySupervisedInstanceSegmentation2022a} and our Simplified PointRend architectures. }
    \label{fig:pointrends}
\end{figure}

\cref{fig:pointrends} illustrates the detailed structure of our Simplified PointRend module compared to Implicit PointRend~\cite{chengPointlySupervisedInstanceSegmentation2022a}. 
Simplified PointRend retains the overall structure of Implicit PointRend, but the dynamic point head is replaced by a common 3-layer perceptron with ReLU as the activation function. During the process of point annotation generation, we use $\rho=10$, $N_{pos}=32, N_{neg}=128, N_{body}=N_{flow}=256$, $\alpha=19$ and $\beta=0.8, \gamma=0.2$. The loss weights are set to $\lambda_{joint}=0.025$ and $\lambda_{body}=1$. 
Both training stages use Adam as the optimizer with a learning rate $1e-2$, and the batch size is set to 128. 

\subsection{\schname{}}
We select Adam as the optimizer and CosineAnnealing as the learning rate scheduler with a maximum learning rate 0.2. The batch size is set to 128. 

\section{More Quantitative Results}
In this section, we presents more experimental results to further validate the effectiveness of our \modname{} and \netname{}.

\subsection{Results of the ``Limb'' Variant of \modname{}}
Our \modname{} has two variants, ``joint'' and ``limb''. While previous discussions focus on the ``joint'' variant, this section presents the performance of \schname{} based on \modname{} of the ``limb'' variant. We compare it to the limb stream of PoseConv3D~\cite{duanRevisitingSkeletonbasedAction2022a}. \cref{tab:exp_lbf} shows that \schname{} outperforms PoseConv3D significantly in all settings on NTU~\cite{shahroudyNTURGBLarge2016} and NTU120~\cite{liuNTURGB1202020a}. 

\subsection{Comparison on Different Action Categories}
\begin{figure}[h]
    \centering
    \includegraphics[width=\linewidth]{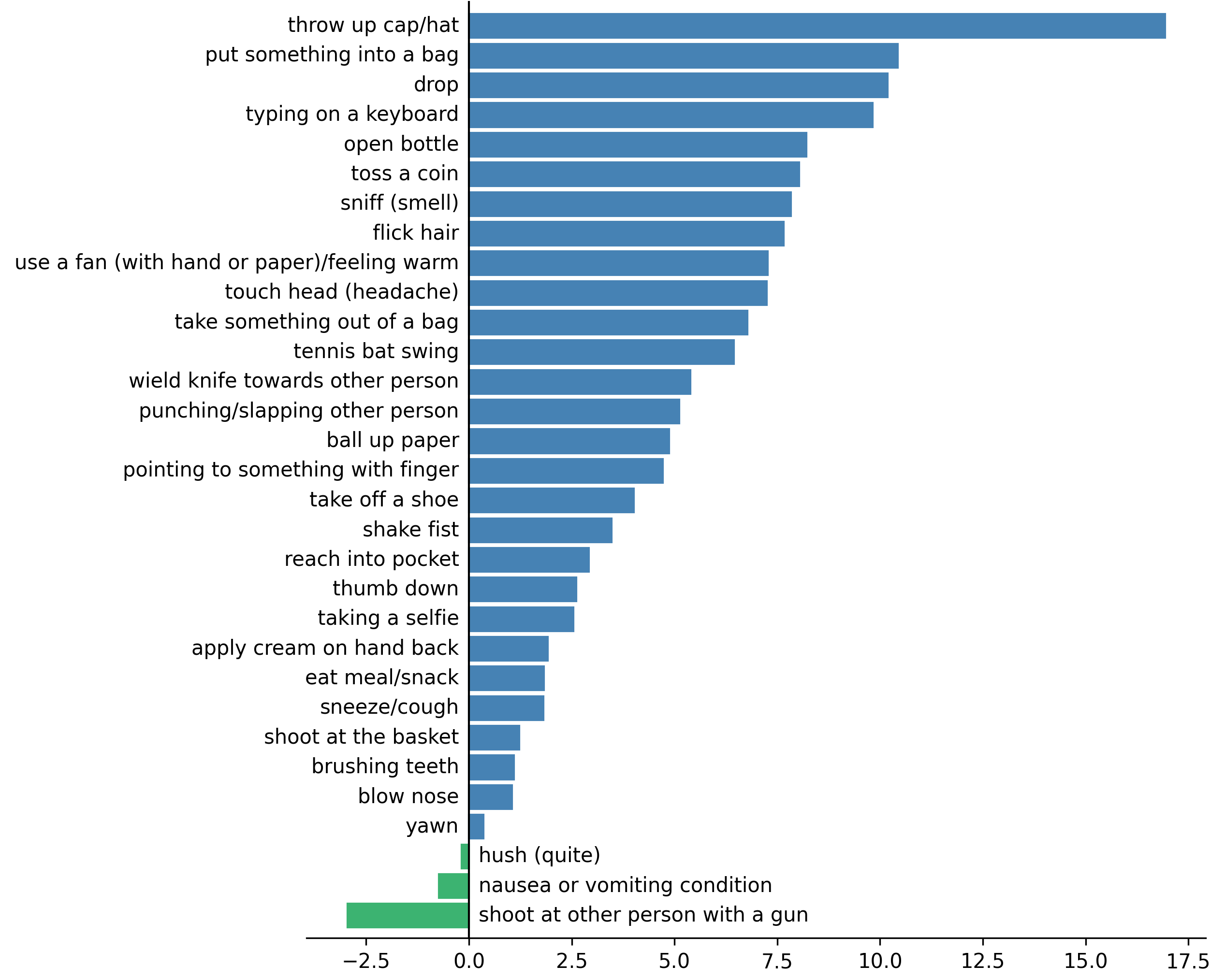}
    \caption{The accuracy difference (\%) between our \schname{} and PoseConv3D for ``Medium'' actions on NTU120 X-Sub.}
    \label{fig:all_actions}
\end{figure}

\cref{fig:all_actions} lists the action-specific accuracy difference between \schname{} and PoseConv3D~\cite{duanRevisitingSkeletonbasedAction2022a} in action categories with the ``Medium'' difficulty level from the NTU120 X-Sub setting~\cite{liuNTURGB1202020a}. \schname{} achieves higher or equal accuracy compared to PoseConv3D in 28 of 31 ``Medium'' actions. Besides, \schname{} outperforms PoseConv3D in 63 out of 74 ``Easy'' actions and in 105 out of 120 total actions, with a maximum improvement of 20.9\%.

\subsection{Comparison with 3D Skeleton}
Our \modname{} captures depth information by using the scale of each joint. One might question why we do not use 3D skeletons, which inherently include depth information. To address this, we evaluate CTR-GCN~\cite{chenChannelwiseTopologyRefinement2021a} using either ground-truth 3D skeletons (GT 3D Skl) or 3D skeletons predicted by a state-of-the-art 2D-to-3D lifting method (Pred 3D Skl). As shown in~\cref{tab:exp_3d}, using ground-truth 3D skeletons results in reduced performance compared to 2D skeletons, due to the inferior quality of 3D data, as noted in~\cite{duanRevisitingSkeletonbasedAction2022a}. Predicted 3D skeletons further degrade accuracy because joint depth cannot be accurately inferred from flat 2D skeletons. This validate that using joint scale to capture depth information is more reliable than directly predicting 3D skeleton. 

\subsection{Ablation Studies}

\begin{description}[style=unboxed, leftmargin=0cm]
\item[Effects of \modname{} Components:] We validate the effectiveness of all three components of \modname{}, scale map volume $\mathcal{S}$, body map $\mathcal{B}$ and flow map $\mathcal{F}$, on NTU X-Sub. Results in~\cref{tab:abl_comp2} demonstrate that all components collectively enhance the performance of our skeleton+\modname{}-based HAR. Notably, the flow map contributes the most, underscoring the importance of incorporating human-object interaction into the HAR pipeline.

\item[Effects of $\mu$ in the joint map volume.] The parameter $\mu$ mentioned in~\cref{sec:jbm_def} balances the skeleton and scale map volume $\mathcal{S}$. According to the results in~\cref{tab:abl_mu}, \modname{} achieves optimal performance when $\mu=0.1$ among the choices. This result aligns with our expectation that the depth information in $\mathcal{S}$ improves the HAR accuracy in some challenging scenarios where using only skeleton can lead to ambiguity.

\item[Effects of Clip Length:] In previous experiments, GCN-based methods process 100 frames per clip, whereas PoseConv3D and \schname{} uses only 48 frames. This allows GCN-based methods to leverage more temporal information, reducing the ambiguity in action understanding. To fairly compare \schname{} and GCN-based methods using the same temporal information, we evaluate CTR-GCN~\cite{chenChannelwiseTopologyRefinement2021a} with 48 frames. Results in~\cref{tab:exp_cl} demonstrate that when the clip length is 48 for all methods, our \schname{} achieves the highest accuracy. Notably, although \schname{} obtains lower performance than CTR-GCN on NTU X-View with a smaller clip length, it outperforms CTR-GCN when both methods use the same clip length. 

\end{description}

\section{More Visualization Results}
This section presents more visualizations of our predicted \modname{} in various datasets. 

\begin{figure}[h]
    \centering
    \includegraphics[width=\linewidth]{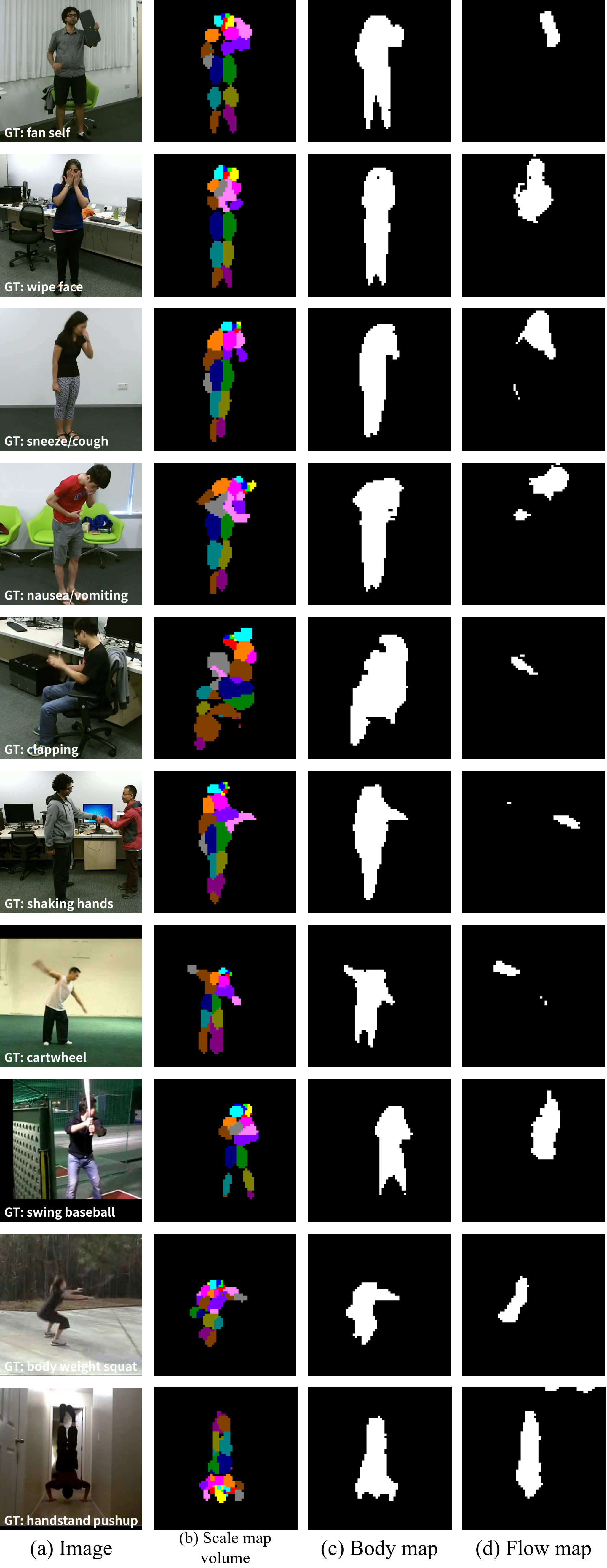}
    \caption{Visualization of \modname{} components predicted by \netname{} on NTU120~\cite{liuNTURGB1202020a} (row 1-6), HMDB51~\cite{kuehneHMDBLargeVideo2011} (row 7-8) and UCF101~\cite{soomroUCF101Dataset1012012} (row 9-10). Each joint is depicted in a distinct color.}
    \label{fig:vis_suppl}
\end{figure}

\end{document}